\documentclass[conference]{IEEEtran}
\usepackage{cite}
\usepackage{amsmath,amssymb,amsfonts}

\usepackage{algorithmic}
\usepackage{graphicx}
\usepackage{textcomp}
\usepackage[dvipsnames]{xcolor}
\usepackage[hidelinks]{hyperref}
\usepackage{mathtools}
\usepackage{multirow}
\usepackage[algo2e,ruled,vlined, noend, linesnumbered]{algorithm2e}
\usepackage{amsthm}
\theoremstyle{definition}
\newtheorem{definition}{Definition}
\usepackage{xparse}
\usepackage{etoolbox}
\usepackage{booktabs}
\usepackage{url}

\makeatletter
\let\MYcaption\@makecaption
\makeatother

\usepackage[font=footnotesize]{subcaption}

\makeatletter
\let\@makecaption\MYcaption
\makeatother

\usepackage{pythonhighlight}
\usepackage{listings}
\definecolor{codegreen}{rgb}{0,0.6,0}
\definecolor{codegray}{rgb}{0.5,0.5,0.5}
\definecolor{codepurple}{rgb}{0.58,0,0.82}
\definecolor{backcolour}{rgb}{0.95,0.95,0.92}

\lstdefinestyle{mystyle}{
    emph={KMeans, SpectralClustering, Agglomerative Clustering, GaussianMixture},
    emphstyle={\color{black}\bfseries},
    commentstyle=\color{codegreen},
    keywordstyle=\color{magenta}\textbf,
    numberstyle=\tiny\color{codegray},
    stringstyle=\color{codepurple},
    basicstyle=\footnotesize\ttfamily,
    breakatwhitespace=false,
    breaklines=true,
    captionpos=b,
    keepspaces=true,
    showspaces=false,
    showstringspaces=false,
    showtabs=false,
    tabsize=2
}
 
\makeatletter
\patchcmd\algocf@Vline{\vrule}{\vrule \kern-0.52pt}{}{}
\patchcmd\algocf@Vsline{\vrule}{\vrule \kern-0.52pt}{}{}
\makeatother

\newcommand{\algname}{DECCS}
\newcommand{\etal}{~\textit{et al.}}
\newcommand{\ensemble}{\mathcal{E}}
\newcommand{\data}{\mathbf{X}}
\newcommand{\enc}{\operatorname{enc}_\Theta}
\newcommand{\dec}{\operatorname{dec}}
\newcommand{\enccons}{\operatorname{enc}_{\Theta_{cr}}}

\newcommand{\agreementT}{\tau}

\newcommand{\embedding}{\mathbf{Z}}
\newcommand{\consrep}{\mathbf{Z}_{cr}}

\newcommand{\ce}{\mathcal{L}^i_{CE}}

\usepackage[acronym,nonumberlist,nogroupskip]{glossaries}  
\glsdisablehyper

\newacronym{cc}{CC}{consensus clustering}
\newacronym{cr}{CR}{consensus representation}
\newacronym{ca}{CA}{co-association}
\newacronym{dc}{DC}{deep clustering}
\newacronym{ae}{AE}{autoencoder}
\newacronym{cf}{CF}{consensus function}
\newacronym{js}{JS}{Johnson–Lindenstrauss}
\newacronym{em}{EM}{Expectation Maximization}
\newacronym{rpfcm}{RP+FCM}{Random Projection Fuzzy c-Means}
\newacronym{fcm}{FCM}{Fuzzy c-Means}
\newacronym{cspa}{CSPA}{Cluster-based Similarity Partitioning Algorithm}
\newacronym{hgpa}{HGPA}{HyperGraph Partitioning Algorithm}
\newacronym{mcla}{MCLA}{Meta-CLustering Algorithm}
\newacronym{nmf}{NMF}{Nonnegative Matrix Factorization}
\newacronym{hbgf}{HBGF}{Hybrid Bipartite Graph Formulation}
\newacronym{ssl}{SSL}{semi-supervised learning}
\newacronym{anmi}{ANMI}{average pairwise normalized mutual information}
\newacronym{nmi}{NMI}{normalized mutual information}
\newacronym{ari}{ARI}{adjusted rand index}
\newacronym{lwea}{LWEA}{Locally Weighted Evidence Accumulation}
\newacronym{eac}{EAC}{Evidence Accumulation}
\newacronym{rpfern}{RP+EM}{Random Projection Expectation Minimization}

\begin{document}

\title{Deep Clustering With Consensus Representations}

\author{\IEEEauthorblockN{Lukas Miklautz\textsuperscript{1,2}, Martin Teuffenbach\textsuperscript{1,2}, Pascal Weber\textsuperscript{1,2}, Rona Perjuci\textsuperscript{1},\\ Walid Durani\textsuperscript{3}, Christian Böhm\textsuperscript{1}, Claudia Plant\textsuperscript{1,4}}
\IEEEauthorblockA{\textsuperscript{1}\textit{Faculty of Computer Science},
\textit{University of Vienna},
Vienna, Austria\\
\textsuperscript{2}\textit{UniVie Doctoral School Computer Science}, Vienna, Austria\\
\textsuperscript{3}\textit{Institute of Informatics, Ludwig Maximilian University of Munich}, Munich, Germany\\
\textsuperscript{4}\textit{ds:UniVie}, Vienna, Austria\\
\{firstname.lastname\}@univie.ac.at, durani@dbs.ifi.lmu.de}
}

\maketitle

\begin{abstract}
The field of deep clustering combines deep learning and clustering to learn representations that improve both the learned representation and the performance of the considered clustering method. Most existing deep clustering methods are designed for a single clustering method, e.g., $k$-means, spectral clustering, or Gaussian mixture models, but it is well known that no clustering algorithm works best in all circumstances. Consensus clustering tries to alleviate the individual weaknesses of clustering algorithms by building a consensus between members of a clustering ensemble. Currently, there is no deep clustering method that can include multiple heterogeneous clustering algorithms in an ensemble to update representations and clusterings together. To close this gap, we introduce the idea of a consensus representation that maximizes the agreement between ensemble members. Further, we propose \algname{} (Deep Embedded Clustering with Consensus representationS), a deep consensus clustering method that learns a consensus representation by enhancing the embedded space to such a degree that all ensemble members agree on a common clustering result.
Our contributions are the following: (1)~We introduce the idea of learning consensus representations for heterogeneous clusterings, a novel notion to approach consensus clustering. (2)~We propose \algname{}, the first deep clustering method that jointly improves the representation and clustering results of multiple heterogeneous clustering algorithms. (3)~We show in experiments that learning a consensus representation with \algname{} is outperforming several relevant baselines from deep clustering and consensus clustering. \\
\end{abstract}


\section{Introduction}

Clustering is the task of unsupervised classification, where we infer cluster labels from the data.\footnote{Our code is available at \url{https://gitlab.cs.univie.ac.at/lukas/deccs}.} \Gls{dc} combines unsupervised deep learning and clustering to learn representations (embeddings) that improve clustering performance. Current \gls{dc} methods are designed with only a single clustering model in mind, e.g., DEC \cite{DEC} which improves the representation for $k$-means \cite{KM_Lloyd82}, VaDE \cite{Vade} for Gaussian mixture models \cite{EM}, DeepECT \cite{ECT} for hierarchical clustering \cite{survey_hierarchical_clustering_MurtaghC12}, and SpectralNet \cite{SpectralNet18} for spectral clustering \cite{spectral_clustering_Luxburg07}. Relying on the assumptions of a single clustering model leads to poor results if the assumptions are not met by the data. 

\Gls{cc} can alleviate the limitations of individual clusterings by combining a clustering ensemble into a single robust clustering \cite{strehl_cluster_2002}. Unfortunately, applying current \gls{cc} methods to high-dimensional data sets leads to unsatisfactory results, because they are either limited to linear transformations \cite{fern_random_2003, popescu_random_2015, bezdek_random_2016}, only work for $k$-means like clusterings \cite{RegattiConsensus21}, or only use \gls{cc} information as input features for \gls{dc} without updating the \gls{cc} in response to improved data representations \cite{iec_LiuSLF16,agae_TaoLLW019}. 

\begin{figure}[t]
\centering
\begin{subfigure}[c]{.9\linewidth}
	\includegraphics[width=\linewidth]{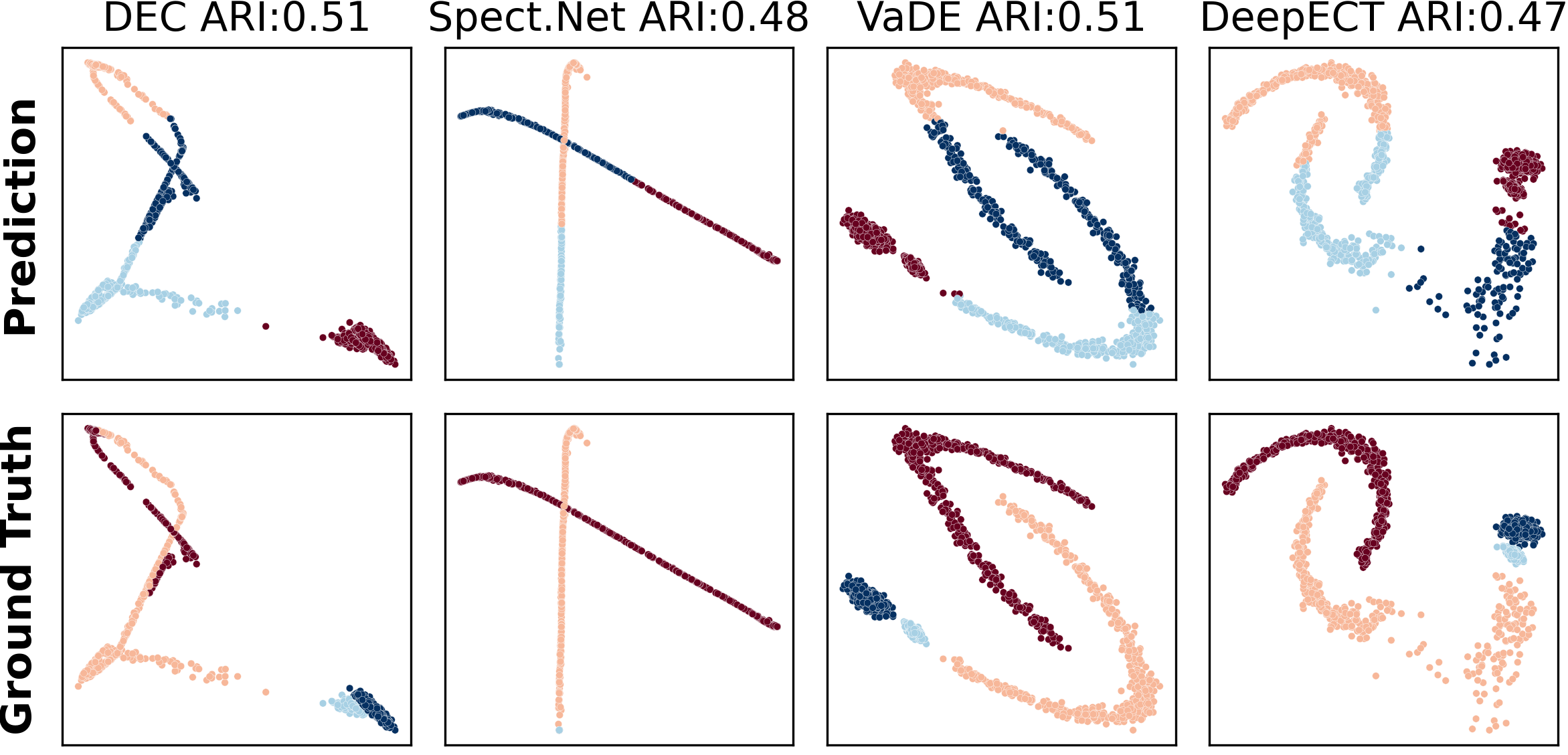}
    \label{fig:deep_clustering}
\end{subfigure}

\caption{Existing \gls{dc} methods are limited by their assumed cluster model. Here they fail, because the data contains clusters of differing shapes.}%
\label{fig:deep_clustering_comparison}%
\end{figure}

In contrast to that, we propose our novel \textbf{D}eep \textbf{E}mbedded \textbf{C}lustering with \textbf{C}onsensus representation\textbf{S} (\algname{}) method, which is a \gls{dc} method that can be applied to high-dimensional data, finds non-linearly hidden clusters and works with many existing clustering algorithms. \algname{} learns a \textit{consensus representation} (CR) that maximizes the agreement between ensemble members.  
The key idea we use for consensus representation learning with \algname{} is that most clustering methods can find well-separated clusters in a low-dimensional space that have a simple shape, e.g., dense, spherical clusters. Using this idea, \algname{} learns a consensus representation by transforming the embedded space such that it is trivial to cluster and, therefore, all ensemble members naturally agree on one partitioning into clusters. %
\begin{figure}[t]
\begin{subfigure}[c]{.2\linewidth}
	\includegraphics[width=\linewidth]{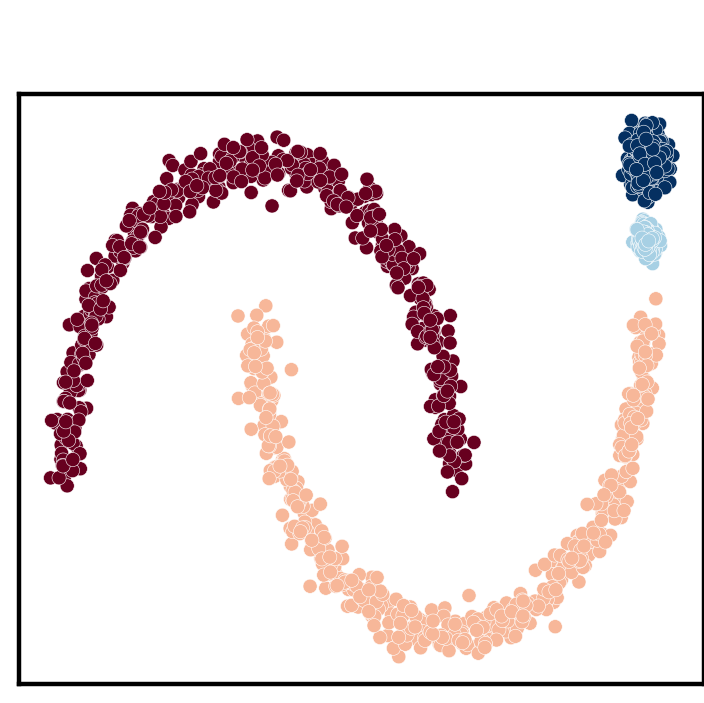}%
	\caption{Original}%
    \label{fig:synthetic_dataset}%
\end{subfigure}%
\hfill%
\begin{subfigure}[c]{.2\linewidth}%
	\includegraphics[width=\linewidth]{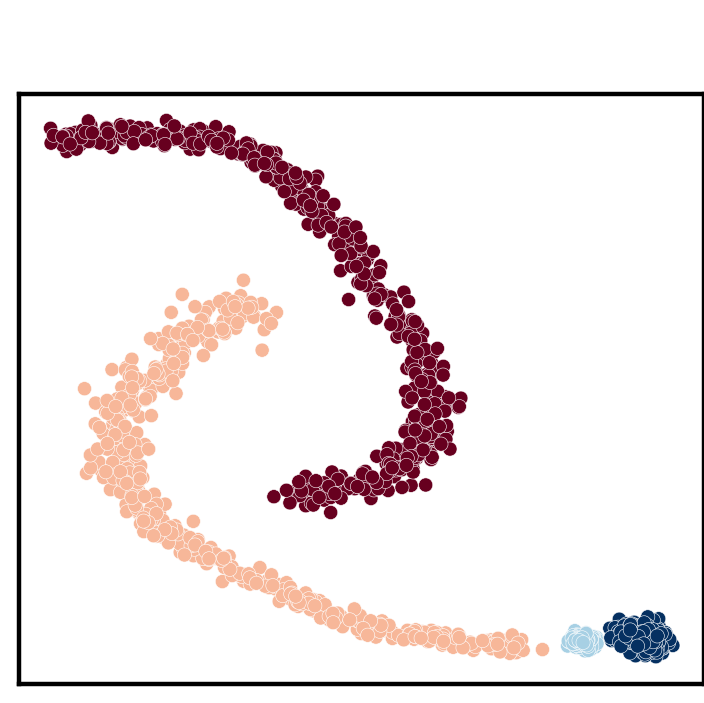}%
	\caption{Initial AE}%
\end{subfigure}%
\hfill%
\begin{subfigure}[c]{.2\linewidth}%
	\includegraphics[width=\linewidth]{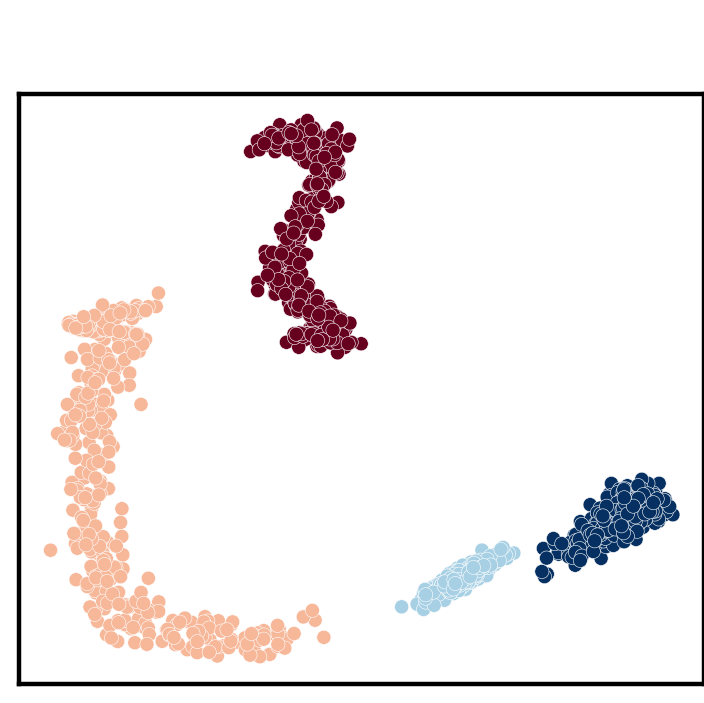}%
	\caption{Update}%
\end{subfigure}%
\hfill%
\begin{subfigure}[c]{.2\linewidth}%
	\includegraphics[width=\linewidth]{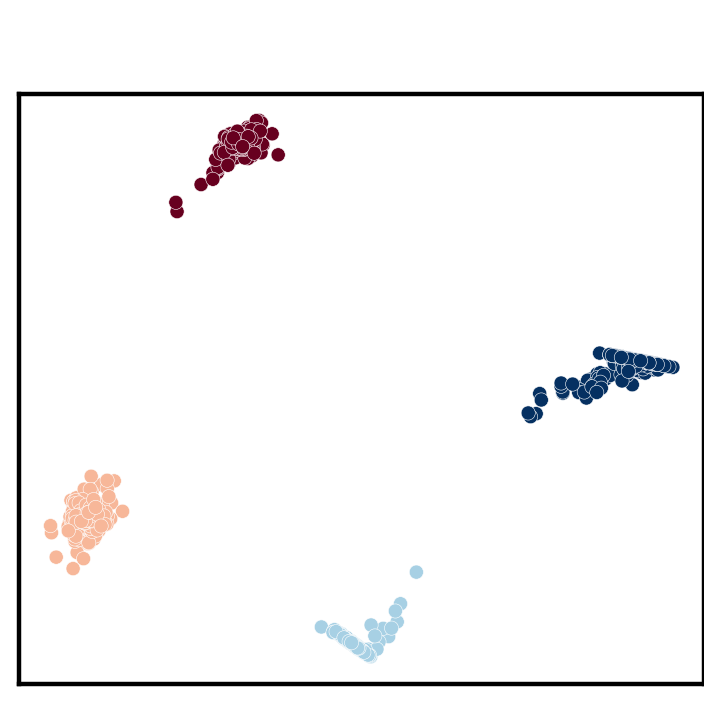}%
	\caption{Final CR}%
\end{subfigure}%
\caption{A synthetic data set \textbf{(a)} containing four clusters is embedded \textbf{(b)} with an \glsfirst{ae}. \algname{} transforms the initial \gls{ae} embedding via several updates \textbf{(c)} to the final \glsfirst{cr} in which clusters are compact and well separated \textbf{(d)}.}%
\label{fig:first_figure}%
\end{figure}%
\begin{figure}
\centering
    \begin{subfigure}[c]{0.85\linewidth}
	\includegraphics[width=\linewidth]{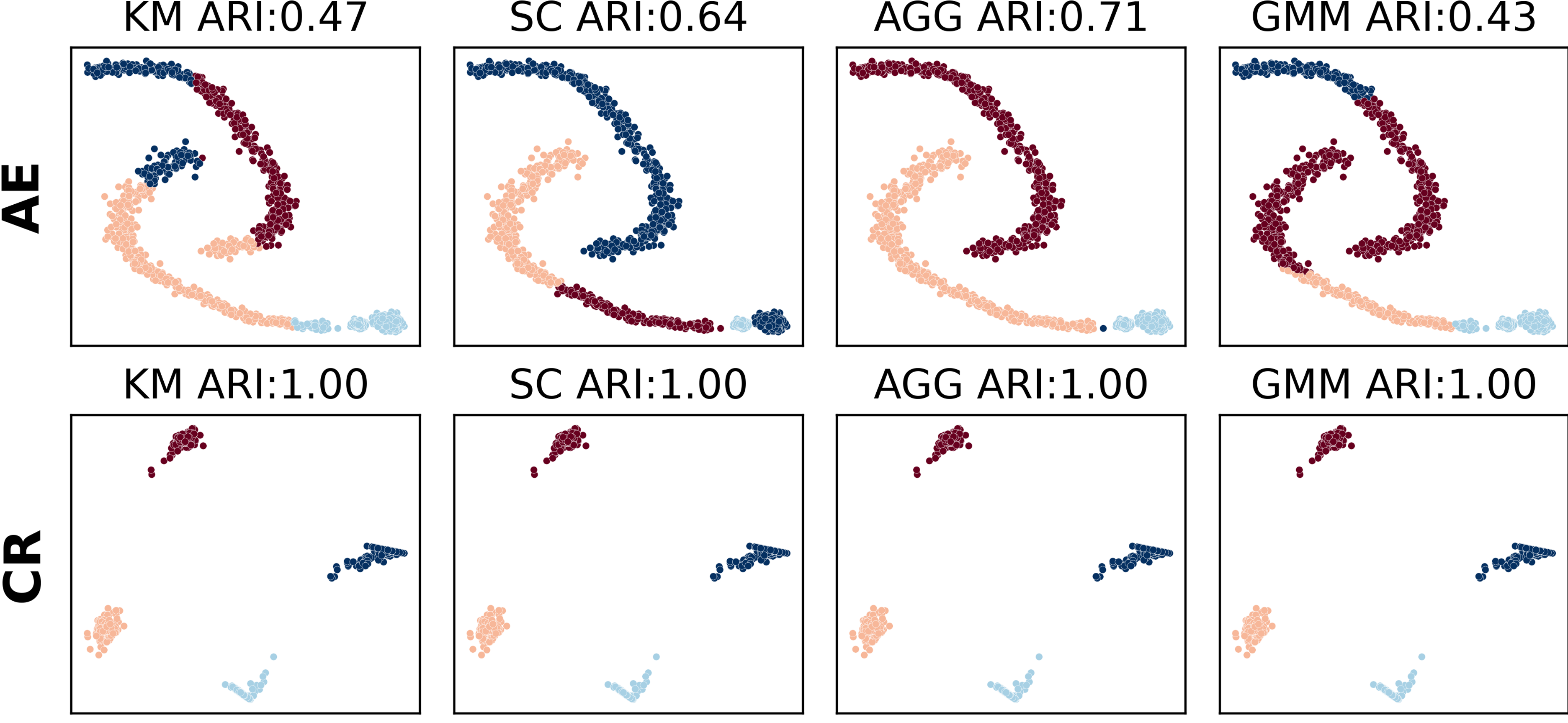}
	\end{subfigure}
	\caption{Cluster performance on initial \gls{ae} embedding and learned CR for an ensemble of $k$-means (KM), Spectral clustering (SC), Agglomerative clustering (AGG), and Gaussian mixture model (GMM). 
	}
    \label{fig:first_figure_ae_cr}
\end{figure}
Fig. \ref{fig:first_figure} illustrates on a synthetic data set how \algname{} transforms an initial embedding that contains clusters of different shapes to a consensus representation that consists only of dense, spherical, and well separated clusters. In Fig. \ref{fig:first_figure_ae_cr}, multiple, heterogeneous algorithms with different assumptions about the cluster structure are applied to the initial \glsfirst{ae} embedding (upper row). Initially, the clustering algorithms perform poorly, but applied to the consensus representation learned with \algname{} all algorithms in the ensemble reach the same, perfect clustering (bottom row) as measured with the \glsfirst{ari} \cite{ARI}.
In Fig. \ref{fig:deep_clustering_comparison}, we apply existing \gls{dc} methods to the same synthetic data set using the same \gls{ae} and plot their learned embeddings. While this data set can be clustered by classical clustering techniques, we see that \gls{dc} methods fail, because their assumptions are not met. For example, DEC is performing poorly because the data set contains non-spherical clusters, which is not suited for $k$-means. As a consequence DEC is producing a distorted embedding (first column).

In this work, we tackle the shortcomings of existing \gls{dc} and \gls{cc} techniques and present the following contributions:
\begin{itemize}
    \item[\textbf{(1)}] We introduce the idea of a consensus representation, which is a representation that maximizes the agreement of the applied clustering algorithms by producing similar clustering results for all clustering methods included in the ensemble.
    \item[\textbf{(2)}] We propose \algname{}, the first \gls{dc} algorithm that can include multiple heterogeneous clustering methods to jointly improve the learned embedding and clustering results by simplifying the representation.
    \item[\textbf{(3)}] Our method is outperforming several relevant baselines in terms of cluster performance.
\end{itemize}

\section{Background - Consensus Clustering}
\label{sec:bg_consensus_clustering}

\Gls{cc} can overcome the limitations of individual clusterings by combining multiple clustering solutions into a single robust partitioning \cite{strehl_cluster_2002}.
In general, \gls{cc} algorithms consist of two stages:
\begin{enumerate}
    \item Generate a set of base partitions using single clustering algorithms (e.g., $k$-means, Spectral Clustering, etc.)
    \item Combine the base partitions using a consensus function to obtain a final partition. 
\end{enumerate}
Traditionally, the two stages are independent of each other. The consensus function does not access the original features of the data set to find the optimal combination of base partitions. 

During the design of the consensus function the goal is to combine a set \textbf{$\Pi$} of $|\Pi|$ partitions $\pi_i$ into one final clustering $\pi_{cc}$, such that $\pi_{cc}$ agrees as much as possible with the base partitions. 
In their framework, \cite{strehl_cluster_2002} suggested to use the \gls{anmi} between the \gls{cc} and the base clusterings as an objective function to measure the agreement: 
\begin{equation}
    \label{eq:strehlobjective}
    \pi_{cc} = \text{argmax}_{\bar{\pi}} \sum_{i=1}^{|\Pi|} \text{NMI}(\pi_i, \bar{\pi})
\end{equation}
Using the \gls{nmi} has the benefit that it is invariant to the permutation and absolute values of cluster labels and allows for a different number of clusters $k_i$ in each  partition $\pi_i$. Further, the \gls{nmi} is symmetric and is 1 if two clusterings match perfectly and 0 if they are independent of each other. 

Instead of the need to design a consensus function to optimize Eq. \ref{eq:strehlobjective} our \algname{} algorithm learns a (non-linear) consensus function to learn the \textit{consensus representation} as we explain in the following. 

\section{Objective Function for Consensus Representation Learning}
\label{sec:problem_setup}

For our novel problem setting, we use an encoder $\enc$ that maps a data point $\mathbf{x} \in \mathbb{R}^{D}$ to a typically lower-dimensional embedded vector $\mathbf{z} \in \mathbb{R}^{d}$, where $\Theta$ are the learnable parameters of the encoder. Then, let $\data$ be an $N\times D$ dimensional input data matrix and $\embedding=\enc(\data)$ be an $N\times d$ dimensional embedded data matrix with $d<D$. 
Further, let  $\ensemble$ be a set of heterogeneous clustering algorithms with potentially different number of clusters $k_i$, where each $i^\text{th}$ member $e_i$ produces a clustering result $\mathbf{\pi}_i=e_i(\embedding)$. We define the consensus representation in the following.

\begin{definition}[Consensus representation $\consrep$]
\label{def:consensus_representation} 
Let $\Theta$, $\enc$, $\data$, $\embedding$, and $\ensemble$ be defined as above. The \textit{consensus representation} $\consrep$ maximizes the following objective function:
\begin{equation}
    \label{eq:min_objective_nmi}
    f_\Theta = c \sum_{i=1}^{|\ensemble|}\sum_{j>i}^{|\ensemble|} \text{NMI}(e_i(\enc(\data)), e_j(\enc(\data)))\text{,}
\end{equation}
with $\consrep:=\enccons(\data)$, where $\enccons$ is the \textit{consensus representation function} and $c$ is a normalization constant $c=\frac{2}{|\ensemble|^2 - |\ensemble|}$ for the equation to sum to one.
\end{definition}

The consensus representation maximizes the agreement of all partitions with each other, where the agreement is measured using the pairwise \gls{nmi} \cite{strehl_cluster_2002}. 
The optimal encoder parameters for the consensus representation $\consrep$ are then learned with 

\begin{equation}
\label{eq:min_objective_nmi_argmax}
\Theta_{cr} = \text{argmax}_{\Theta} f_\Theta \text{.}
\end{equation}

Note that Eq. \ref{eq:min_objective_nmi} allows for degenerate solutions, like setting $\consrep$ to a constant if $\enc$ is non-linear. To avoid degenerate solutions in practice we include regularizers in the objective, like enforcing the invertibility of $\consrep$ back to $\data$ by using the \gls{ae} reconstruction loss. In the following, we introduce our \algname{} method and illustrate how it approaches the consensus representation learning problem.

\section{Method - \algname{}}
\label{sec:method}

We motivate our consensus representation learning approach using the observation that most clustering methods are able to detect compact and well-separated clusters in low-dimensional spaces. \algname{} uses the cluster information from all ensemble members to learn such a simplified representation. Decreasing the ambiguity of the representation during training will increase the similarity of the clusterings that ensemble members will produce, which subsequently increases the pairwise NMI (Eq. \ref{eq:min_objective_nmi}) of clustering results. \algname{} works by alternating between representation and clustering update steps until an agreement is reached through the consensus representation. 
In the following, we explain our approach in more detail.

\subsection{Overview}

We use a (non-linear) \glsfirst{ae} to learn $\operatorname{enc}$ by reconstructing the original input data $\mathbf{x}$ from $\mathbf{z}$ using the decoder $\dec$ resulting in $\mathbf{\hat{x}}:=\operatorname{dec}(\mathbf{\operatorname{enc}(\mathbf{x})})$.\footnote{We reuse $\operatorname{enc}$ here, whether it is vector or matrix-valued should be clear from the context.} The \gls{ae} reconstruction $\mathbf{\hat{x}}$ is learned by minimizing a reconstruction loss $\mathcal{L}_{\text{rec}} = \|\mathbf{x} - \mathbf{\hat{x}}\|$, e.g., using the mean squared error. 
Given the \gls{ae}, our \algname{} algorithm consists of three main steps that we explain in the following sections. First, we illustrate how we generate a set of base partitions by applying the cluster ensemble to a sub-sample of the embedding in Section \ref{sec:base_partitions}. Second, we show how to approximate each partition with a classifier to label the remaining data points in Section \ref{sec:approx_base_partitions}. Third, we state our consensus objective in Section \ref{sec:consensus_objective} and in Section \ref{sec:consensus_representation_learning}, we show how the consensus representation is updated. The algorithm is presented in Section \ref{sec:algorithm}.

\subsection{Generating base partitions}
\label{sec:base_partitions}

At the beginning of each round $t$ of our algorithm, we draw a small random sample $\data_t$ of size $n<N$ from $\data$, because some clustering algorithms are impractical to be applied to large data sets and re-sampling can make the \gls{cc} more robust \cite{bootstrapp_resampling_ensemble}.
Next, we embed the sample using $\enc(\data_t)=\embedding_t$ and generate a set of base partitions $\Pi_t$ by applying all ensemble members to the embedding $\pi_i=e_i(\embedding_t)$. The sampling procedure and the low-dimensional embedded space allow us to use more run-time and memory expensive algorithms, such as spectral clustering, in our ensembles. Further, using the sampling and heterogeneous ensembles, we can achieve a sufficiently diverse set of base partitions $\Pi_t$. 

\subsection{Approximating base partitions}
\label{sec:approx_base_partitions}

Since we only have cluster labels for $n<N$ data points due to the random sampling, but require cluster labels for all N data points, we use a classifier to approximate the clustering for the remaining $N-n$ points. 
We approximate the set of base partitions $\Pi_t$ using a set of classifiers $\mathcal{G}_t$, where classifier $g_i$ is trained to predict the corresponding clustering $\pi_i$. We train each classifier by minimizing the cross-entropy loss of cluster labels $\pi_i$ and its prediction, i.e., \begin{equation}
\label{eq:cross_entropy}
\mathcal{L}_{\text{CE}} = \sum_{i=1}^{|\Pi_t|}\ce=-\frac{1}{n}\sum^n_{j=1}\sum^{k_i}_{l=1}I[l=\pi_{i,j}]\operatorname{log}g_i(\mathbf{z}_j)\text{,}
\end{equation} where $\pi_{i,j}$ is the cluster label corresponding to the $j^\text{th}$ data point and $I$ is the indicator function. 
While in principle one can use any classifier for $g_i$ we chose linear classifiers with the Softmax function as output, i.e., $g_i(\mathbf{x})=\operatorname{softmax}(\mathbf{W}_i\mathbf{x} + \mathbf{b}_i)$ with $\mathbf{W}_i$ and $\mathbf{b}_i$ as weights and bias terms respectively. 
The linear classifiers can be trained with little overhead, having only $d \cdot k_i + k_i$ trainable parameters. Updating the linear classifiers together with the non-linear encoder allows us then to approximate non-linear clusterings as well.

\subsection{Consensus Objective}
\label{sec:consensus_objective}

Optimizing Eq. \ref{eq:min_objective_nmi} from Definition \ref{def:consensus_representation} directly is not possible, because the cluster ensemble members are not differentiable. Thus, we learn a low-dimensional representation in which all clusters are spherical, dense, and well separated, such that the ensemble members trivially agree on one partition. To transform non-spherical-shaped clusters into spherical clusters we "move" cluster points closer to their cluster representatives. We choose the mean center of a cluster as the representative because it is stable across update steps, but other choices like the median are also possible. In the following, we use the terms representative and center interchangeably.

Let $\mathcal{C}^j_{\pi_i}$ be the set of data points in the $j^\text{th}$ cluster of partition $\pi_i$, then we can calculate its center $\mathbf{\mu}_j$ using $\mathbf{\mu}_j = \frac{1}{|\mathcal{C}^j_{\pi_i}|} \sum_{\mathbf{x}\in\mathcal{C}^j_{\pi_i}} \enc(\mathbf{x})\text{,}$ and subsequently can construct the $k_i \times d$ matrix $M_i$ containing $k_i$ centers $\mathbf{\mu}_j$ as row vectors. Next, we define our differentiable consensus objective as
\begin{align}
\label{eq:consensus_loss}
    \mathcal{L}_{\text{cons}} &= \sum^{|\Pi_t|}_{i=1} \mathcal{L}^i_{\text{cons}} = \sum^{|\Pi_t|}_{i=1}\|\mathbf{A}_i\mathbf{M}_i - \enc(\data_t)\|_F^2 \\
    &= \sum^{|\Pi_t|}_{i=1}\sum^{k_i}_{l=1}\sum_{\mathbf{x}\in\mathcal{C}^l_{\pi_i}}\|\mathbf{\mu}_l- \enc(\mathbf{x})\|_2^2\nonumber \text{,}
\end{align}
where $\mathbf{A}_i$ is the $n \times k_i$ one hot encoded cluster assignment matrix of partition $\pi_i$, $\|\cdot\|^2_2$ the squared Euclidean norm, and $\|\cdot\|^2_F$ the squared Frobenius norm. Here $\mathbf{M}_i$ and $\mathbf{A}_i$ are fixed, so the encoder $\enc$ has to learn parameters $\Theta$ that map embedded data points $\mathbf{z}=\enc({\mathbf{x}})$ as close as possible to their assigned centers across all partitions. Note, that data points that are close to similar centers across partitions will receive a higher gradient update due to the summation and are thus gathering faster than data points that have conflicting assignments. The centers alone can not capture complex cluster structures properly, which is why we include the cross-entropy loss (Eq. \ref{eq:cross_entropy}) in our objective, as we explain in the next Section.

\subsection{Updating consensus representation}
\label{sec:consensus_representation_learning}
\begin{figure}
\centering
\begin{subfigure}[b]{.22\linewidth}
	\includegraphics[width=\linewidth]{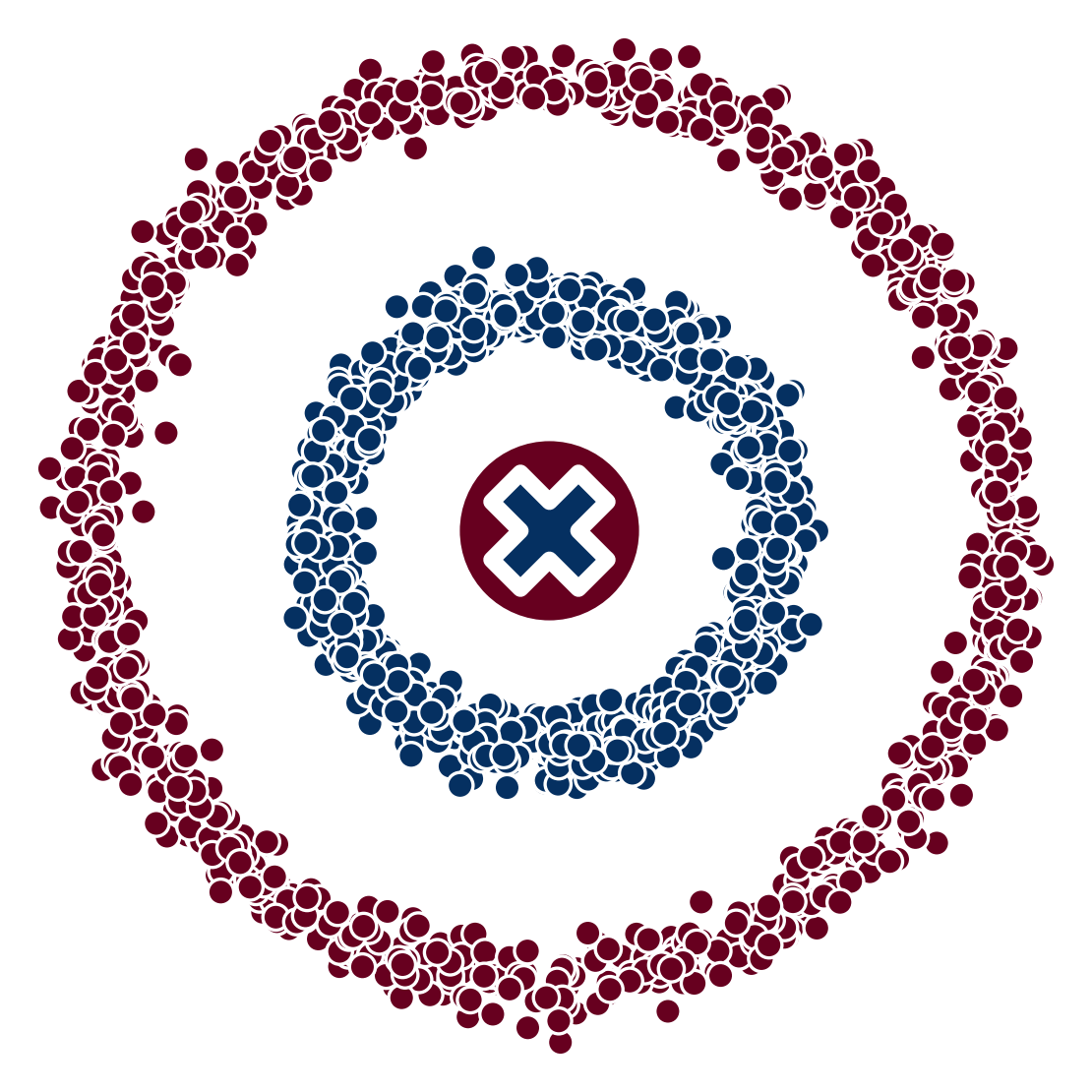}
	\caption{Original}
    \label{fig:consensus_update_toy1}
\end{subfigure}
\hfill
\begin{subfigure}[b]{.22\linewidth}
	\includegraphics[width=\linewidth]{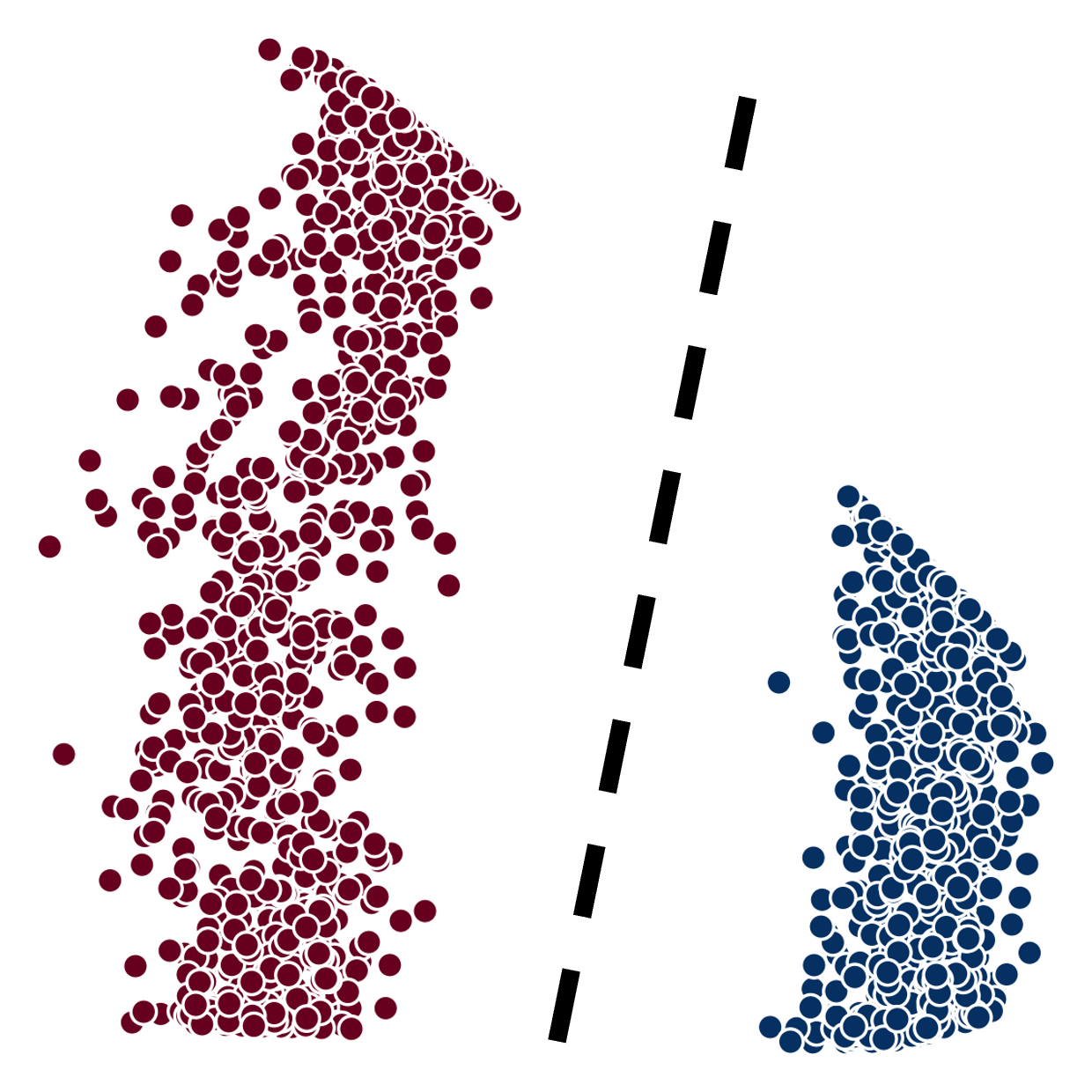}
	\caption{$\mathcal{L}_{\text{CE}}$}
    \label{fig:consensus_update_toy3}
\end{subfigure}
\hfill
\begin{subfigure}[b]{.19\linewidth}
	\includegraphics[width=\linewidth]{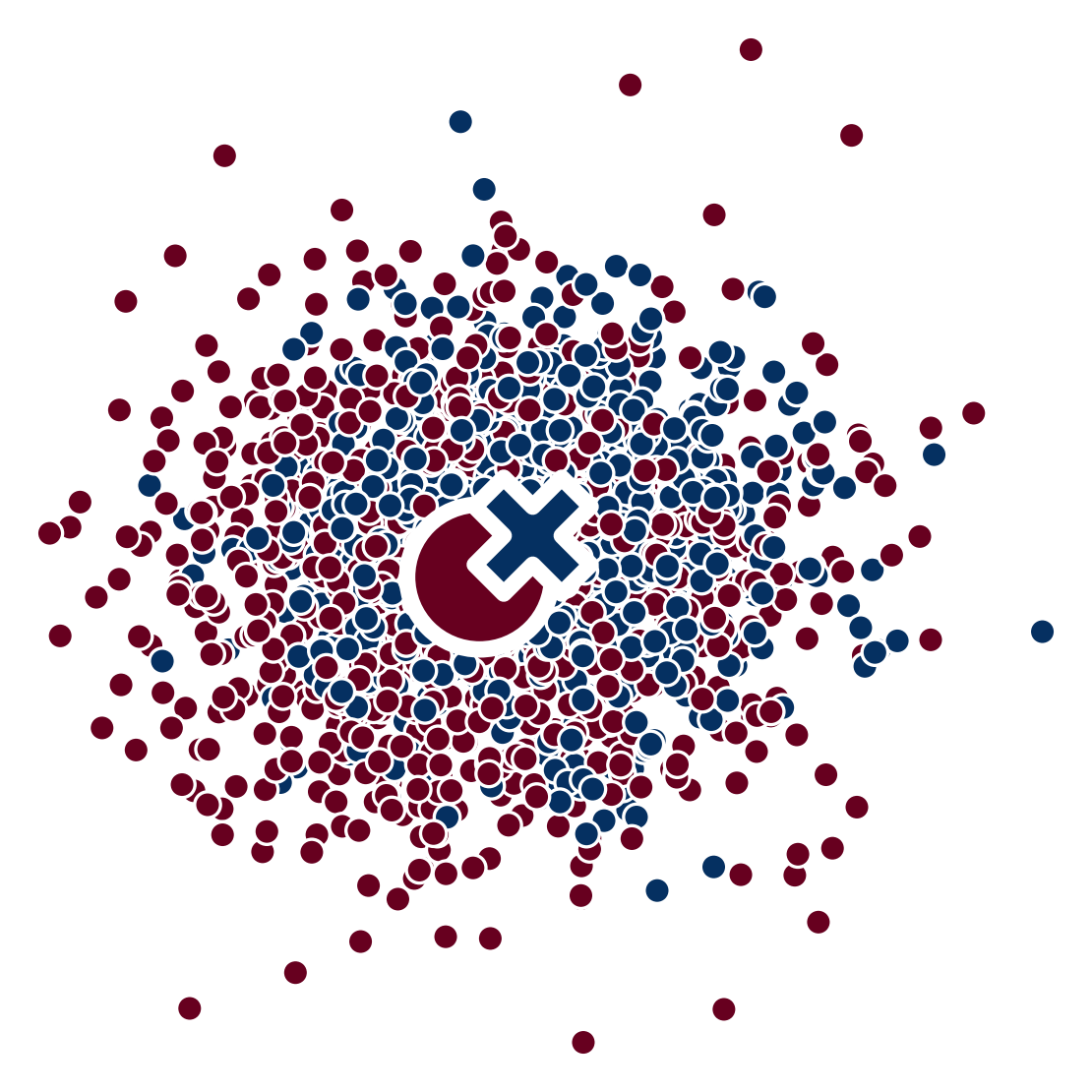}
	\caption{$\mathcal{L}_{\text{cons}}$}
    \label{fig:consensus_update_toy2}
\end{subfigure}
\hfill
\begin{subfigure}[b]{.22\linewidth}
	\includegraphics[width=\linewidth]{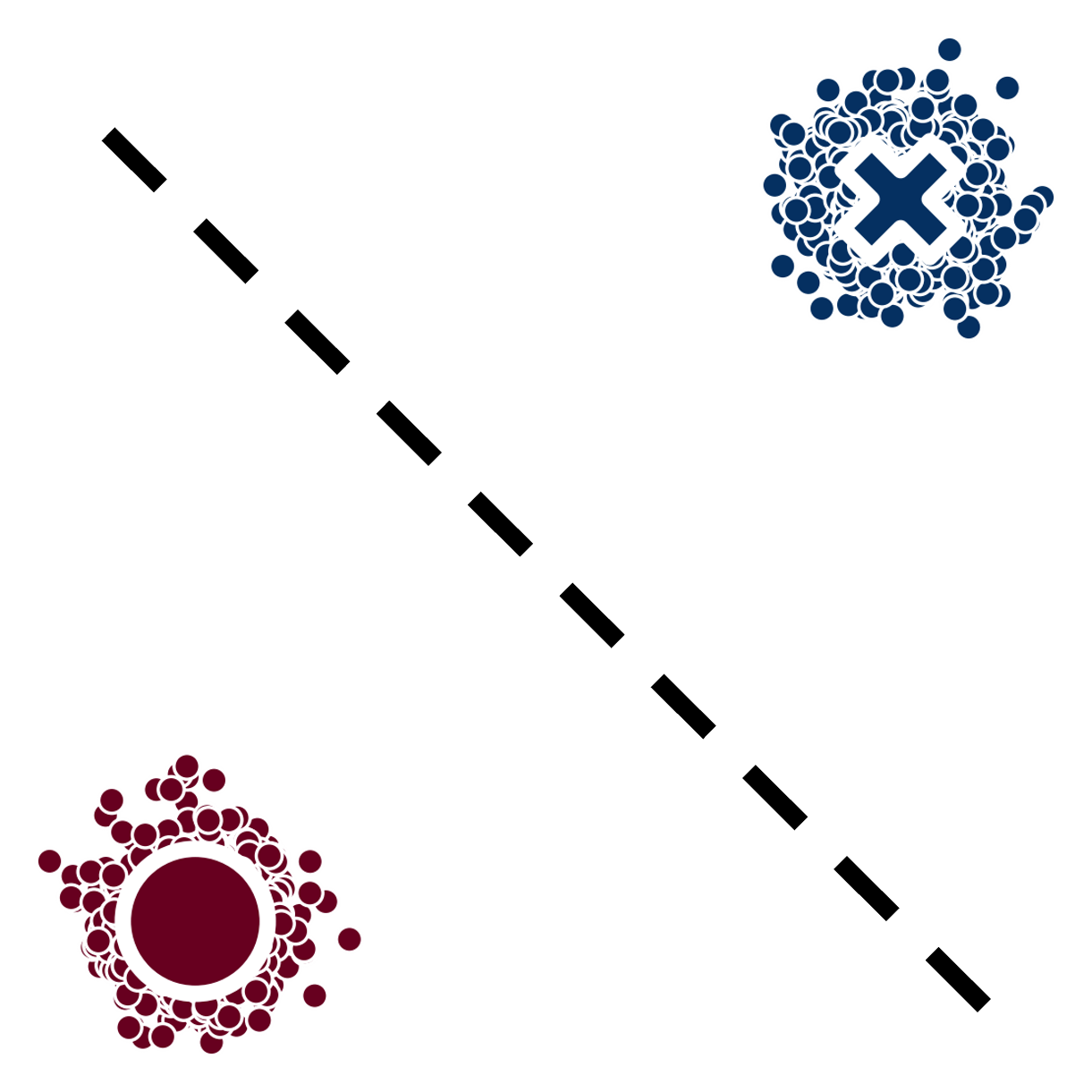}
	\caption{$\mathcal{L}_{\text{CE}}+\mathcal{L}_{\text{cons}}$}
    \label{fig:consensus_update_toy4}
\end{subfigure}
\caption{Example of how optimizing $\mathcal{L}_{\text{CE}}$ together with $ \mathcal{L}_{\text{cons}}$ is able to enhance the representation even if cluster centers overlap. \textbf{(a)} A data set with two clusters (red and blue circle) that have the same centers (red dot and blue cross). \textbf{(b)} Optimizing \algname{} only with $\mathcal{L}_{\text{CE}}$ learns a representation in which both clusters are separated by the decision boundary (dashed black line), but cluster shapes are distorted. \textbf{(c)} Only using $ \mathcal{L}_{\text{cons}}$ merges both clusters. \textbf{(d)} Optimizing $\mathcal{L}_{\text{CE}}$ together with $\mathcal{L}_{\text{cons}}$ separates the clusters and transforms them from circles to spheres, a representation that can be easily clustered.}%
\label{fig:consensus_update_toy}%
\end{figure}%

Putting everything together the objective of our \algname{} algorithm is \begin{equation}\label{eq:method_objective}
    \mathcal{L}=
    \sum_i^{|\Pi_t|}\lambda_i(\ce +  \mathcal{L}^i_{\text{cons}})
    + \lambda_{rec}\mathcal{L}_{\text{rec}}\text{,}
\end{equation}
where $\lambda_i=\frac{1}{|\Pi_t|-1}\sum^{|\Pi_t|}_{j=1} \text{NMI}(\pi_i,\pi_j) $ is a weighting parameter based on the agreement of partition $\pi_i$ with all other partitions measured as average pairwise \gls{nmi}  to exclude random partitions and downscale outlier partitions. We choose the commonly used \gls{ae} reconstruction loss $\mathcal{L}_{\text{rec}}$ as data dependent regularizer to avoid degenerate solutions by keeping $\embedding$ approximately invertible. The hyperparameter $\lambda_{rec}$ weights the importance of $\mathcal{L}_{\text{rec}}$.

The cross-entropy loss $\ce$ of each classifier is included to make sure that the updated representation is still predictive for each partition, e.g., by avoiding the merging of clusters if centers of different clusters are the same, as illustrated in Fig. \ref{fig:consensus_update_toy}. Additionally, we show in the Appendix in Fig. \ref{fig:non-linear-cluster} that this still works even if half of the ensemble members are no better than chance. 

Eq. \ref{eq:method_objective} can be optimized using stochastic gradient descent for a fixed amount of update steps. Once the training has stopped, our algorithm starts again with a new round by applying the clustering algorithms, which will adjust their clustering results to the updated embedding. Then it again pretrains classifiers to approximate them and minimizes $\mathcal{L}$. These steps are repeated for several rounds until a stable agreement is achieved or a maximum number of rounds $T$ has been reached. In the following section, we explain our algorithm in more detail.

\begin{algorithm2e}[ht]
\small
\SetKwInOut{Parameter}{Param}
\SetKwInOut{Input}{Input}
\SetKwInOut{Output}{Output}
\Parameter{Agreement function $a(\cdot)$, agreement threshold $\agreementT$, subsample size $n$, regularization weight $\lambda_{rec}$, rampup function $w(\cdot)$, maximum number of rounds $T$, number of mini-batch training iterations $\text{ITER}$}
\Input{data set $\data$\\initial representation function $\enc$\\ensemble of clustering algorithms $\ensemble$
}
\Output{Estimated consensus representation $\mathbf{\hat{Z}}_{cr}$ and corresponding consensus clustering $\hat{\pi}_{cc}$} 
\BlankLine
$t=0$\;
\While{$t \leq T$}{
    draw sample $\data_t$ of size $n$ from $\data$ and create corresponding embedding  $\embedding_t=\enc(\data_t)$\;
    \tcp{Generate new base partitions}
    generate new empty list for cluster partitions $\Pi_t$\;
    \ForEach{\text{ensemble member} $e_i \in \ensemble$}{
        insert cluster prediction $\pi_i = e_i(\embedding_t)$ into $\Pi_t$\;
    }
    \tcp{Approximate base partitions}
    initialize list of classifiers $\mathcal{G}_t$\;
    \ForEach{\text{cluster prediction} $\pi_i \in \Pi_t$}{
        pretrain classifier $g_i$ by minimizing $\ce(g_i(\embedding_t), \pi_i)$\;
        insert pretrained classifier $g_i$ into $\mathcal{G}_t$\;
    }
    \tcp{Stop if stable agreement or $T$ is reached}
    \If{$(t > 0) \land ((\|a(\Pi_t) - a(\Pi_{t-1})\|_1 < \agreementT) \lor (t==T))$}{
        $\mathbf{\hat{Z}}_{cr}:=\enc(\data)$,  $\hat{\pi}_{cc}:=g_0(\mathbf{\hat{Z}}_{cr})$\;
        break\;
    }
    \Else{
        \tcp{Update consensus representation}
        \While{$j < \text{ITER}$}{
            \ForEach{mini-batch $\mathcal{B} \in X$}{
                calculate for $\mathcal{B}$ loss $\mathcal{L}$:\\\quad$\sum_i^{|\Pi_t|}\lambda_i(\ce +  w(t)\mathcal{L}^i_{\text{cons}}) + \lambda_{rec}\mathcal{L}_{\text{rec}}$\;
                update $\enc$ and $\mathcal{G}_t$ using $\mathcal{L}$\;
                $j = j + 1$\;
            }
            
        }
    }
    $t = t + 1$\;
}
\Return $\mathbf{\hat{Z}}_{cr}$, $\hat{\pi}_{cc}$\;
\caption{\algname{}\label{code:deccs}}
\end{algorithm2e}

\subsection{Algorithm}
\label{sec:algorithm}
\begin{figure*}[t]
  \centering
  \includegraphics[width=0.82\linewidth]{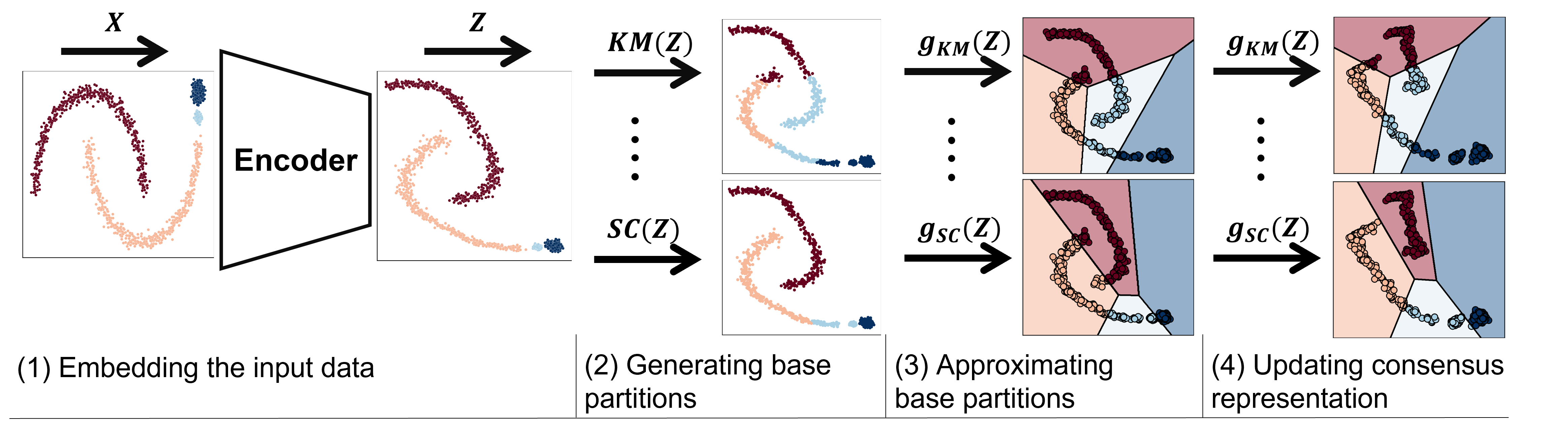}
  \caption{Visualisation of one round of \algname{}. (1) The encoder is used to embed data points $\data$. (2) Clustering results are generated by applying ensemble members $\ensemble=\{\text{KM, \dots, SC}\}$ to $\embedding$. (3) Classifiers $g_i$ are trained to predict the corresponding cluster labels $\pi_i$ from $\embedding$. (4) $\embedding$ is updated via minimizing $\mathcal{L}$.}
    \label{fig:explanation_figure_algorithm}
\end{figure*}
\begin{figure}
  \centering
  \includegraphics[width=0.89\linewidth]{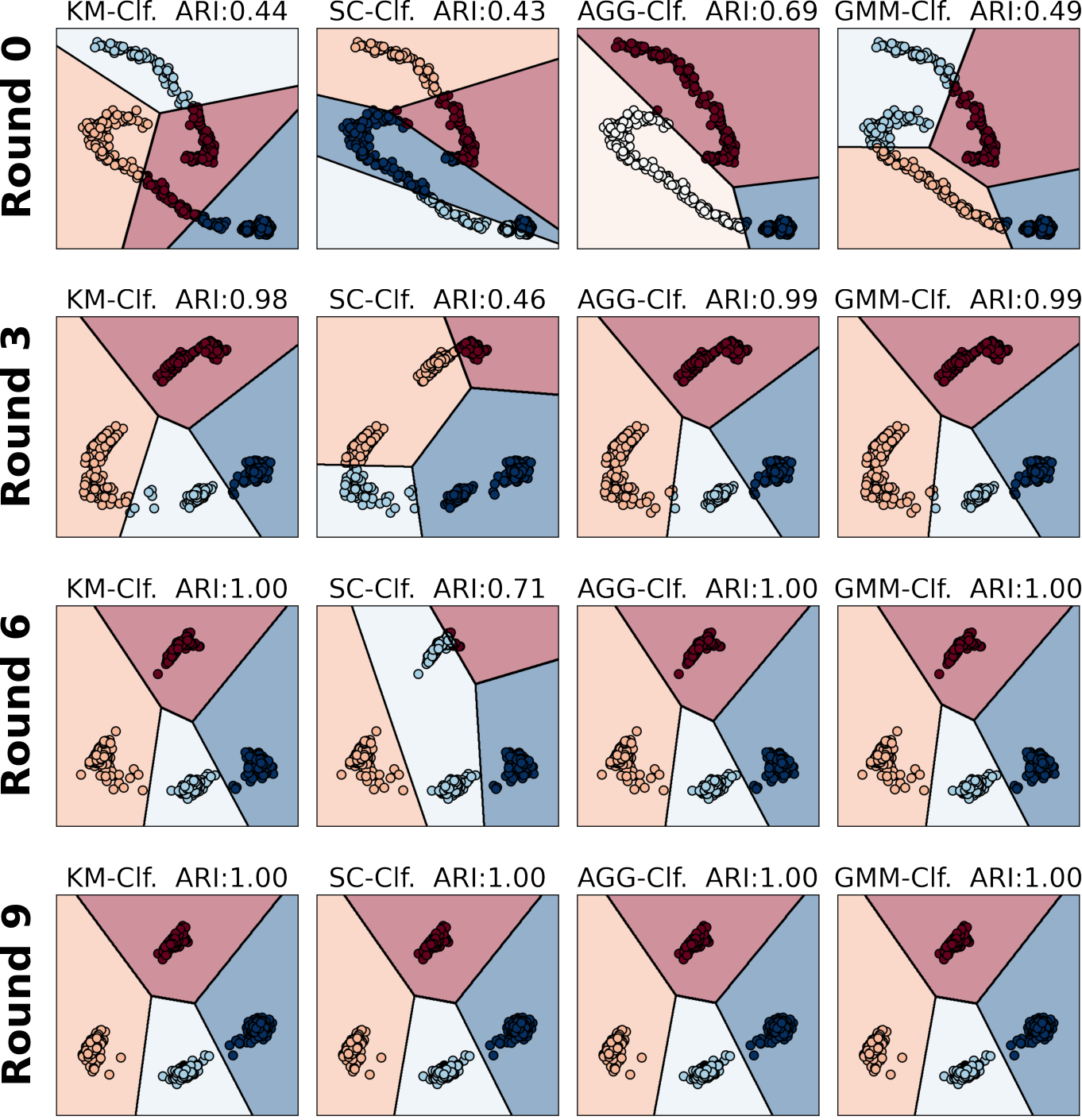}
  \caption{Consensus representation learning with \algname{} over several rounds for the synthetic data set and $\ensemble=\{\text{KM}, \text{SC}, \text{AGG}, \text{GMM}\}$. Each plot shows the classification boundaries for each classifier (Clf.) trained on the cluster partitions. Over several rounds the clusters get better separated and more compact, leading to the same clustering for all ensemble members.}%
    \label{fig:explanation_figure}%
\end{figure}%
Given a data set $\data$, the pretrained encoder $\enc$, and a parameterized ensemble of clustering methods $\ensemble$, we learn a consensus representation $\consrep$ and subsequently a consensus clustering $\pi_{cc}$ with our \algname{} algorithm in the following way. We encode the input data using $\enc$, generate the base partitions by applying cluster ensemble members on $\embedding_t$, approximate the base partitions using classifiers $g_i$ and update the representation by minimizing $\mathcal{L}$. We repeat these steps for several rounds until a stable agreement is achieved or we reached a maximum number of rounds $T$. As agreement function $a(\Pi_t)$ we use the average pairwise \gls{nmi}  between all clusterings in the set of partitions $\Pi_t$. We measure the stability of the agreement by calculating $\|a(\Pi_t) - a(\Pi_{t-1})\|_1 < \tau$, where $\tau$ is the cluster agreement threshold, a user-specified parameter, and $\|\cdot\|_1$ is the absolute distance between the agreement of two subsequent sets of partitions. After the algorithm stops, it returns the estimated consensus representation $\mathbf{\hat{Z}}_{cr}$ and it's corresponding estimated consensus clustering $\hat{\pi}_{cc}$. The consensus clustering $\hat{\pi}_{cc}$ is obtained by applying a clustering algorithm from the ensemble, e.g., $k$-means with the desired $k$ to $\mathbf{\hat{Z}}_{cr}$. If the number of clusters is the same in all ensemble members ($k_i=k_j,  \forall e_i, e_j \in \ensemble$), we choose for $\hat{\pi}_{cc}$ just the result of one of the ensemble members. The pseudocode of our algorithm is depicted in Algorithm \ref{code:deccs}. In Fig. \ref{fig:explanation_figure_algorithm}, we have a visual illustration of one round of our \algname{} algorithm applied to a synthetic data set, and Fig. \ref{fig:explanation_figure} shows its optimization over several rounds.

We use three heuristics for the optimization of \algname{}. First, to speed up convergence we include the predicted cluster labels of the $N-n$ unclustered data points for each classifier $g_i$ during the computation of $\mathcal{L}^i_{\text{cons}}$. These predictions are updated during each mini-batch iteration for unclustered data points in the mini-batch $\mathcal{B}$. Second, to account for the classifiers' uncertainty we weight each distance computation in $\mathcal{L}^i_{\text{cons}}$ with $\alpha_{i,l}=g_{i,l}(\enc(\mathbf{x}))$, which is the $l^\text{th}$ entry of the prediction probability vector of classifier $g_i$.
Third, to enforce the consensus over time $t$, we increase the weight of the consensus loss until a maximum weight $\lambda_{cons}$ is reached. We use the sigmoid schedule as rampup function $w$, like \cite{sigmoid_rampup_TarvainenV17}, to increase the weight $w(t)$ from 0 to $\lambda_{cons}$ over time. 
In total, our algorithm needs the following user-specified parameters, an agreement threshold $\tau$ that indicates how small the agreement gap between two subsequent sets of partitions should be. The data sampling size $n$, which should be chosen w.r.t. computational constraints and demands of clustering algorithms to have a sufficient number of samples. The maximum consensus weight $\lambda_{cons}$ is a hyperparameter that together with the regularization weight $\lambda_{rec}$ trades-off the confidence in the chosen ensemble with the structure of the underlying data. The maximum number of rounds T and the maximum number of mini-batch iterations $\text{ITER}$ for the consensus representation update can be set based on computational constraints. We speed up the training of classifiers and encoders using early stopping, a heuristic that stops training once the loss on a held-out evaluation set starts to increase due to overfitting.

\section{Related Work}
\subsection{Consensus Clustering}
Based on the \gls*{cf} consensus clustering methods can be broadly categorized into median partition- and object co-occurrence based methods. Median partition methods find a partition that is most similar to all the base partitions. Object co-occurrence based methods utilize the \gls{ca} matrix to find the ideal partitioning, where the entries of this matrix reflect how often every two instances are partitioned together. Fred \etal\cite{fred_combining_2005} introduced \gls{eac}, a hierarchical clustering algorithm that uses entries of the \gls{ca} matrix as a similarity measure that is used to produce the final clustering. More recently, \cite{huang_locally_2018} extended this idea by proposing \gls{lwea}, introducing an entropy-based weighting schema, which makes it more robust to outlier partitions. 
Strehl\etal\cite{strehl_cluster_2002}, and later Fern\etal\cite{fern_solving_2004}, utilized the \gls{ca} matrix to formulate graph-based algorithms as a consensus function. 
Li\etal\cite{li_solving_2007} proposed a more efficient \gls{nmf} based algorithm to factorize the \gls{ca} matrix as an alternative.

To generate base partitions for high dimensional data, like images, a line of research follows the idea of random projections (RP). Inspired by the \gls{js} lemma \cite{beals_extensions_1984}, \cite{fern_random_2003} introduced with \gls{rpfern} the first RP-based \gls{cc} algorithm, where the data is projected onto various lower-dimensional subspaces using random matrices. The entries of the resulting \gls{ca} matrix are then used for a hierarchical clustering approach. Similar to this idea, \cite{popescu_random_2015} proposed \gls{rpfcm}, where each subspace is clustered with a \gls{fcm} algorithm. Those partitions are then combined with an agreement function. However, contrary to \algname{}, RP methods are limited to linear transformations.

\subsection{Deep Clustering}

Most, current \gls{dc} methods are designed with only a single clustering model in mind, e.g., SpectralNet \cite{SpectralNet18} for spectral clustering, DEC \cite{DEC}, IDEC \cite{IDEC}, DCN \cite{DCN_YangFSH17} for $k$-means like clustering, VaDe \cite{Vade} for Gaussian mixture models, or DeepECT \cite{ECT} for hierarchical clustering or ENRC \cite{ENRC_MiklautzMABP20} for non-redundant clustering, see \cite{deepclustering_survey_aljalbout18} for an overview. SpectralNet is a deep extension of spectral clustering for large data and out-of-sample generalization. DEC minimizes a soft auxiliary target distribution using the Kullback-Leibler divergence, which is related to soft $k$-means \cite{Jabi21}. Improved DEC (IDEC) includes the \gls{ae} reconstruction loss in the DEC objective to avoid arbitrary clustering results. In contrast to the soft clustering objective of DEC, the DCN algorithm uses hard cluster assignments together with an alternating optimization scheme. It alternates between $k$-means clustering and representation update to achieve a $k$-means friendly embedding. VaDE combines a Gaussian mixture model prior with a variational autoencoder (VAE) \cite{VAE} to learn a deep generative clustering. DeepECT \cite{ECT} introduced a deep embedded cluster tree to learn a hierarchical embedding. 

The ConCURL\cite{RegattiConsensus21} algorithm leverages image augmentation and RPs to learn a cluster ensemble of Softmax predictions to improve the overall clustering performance. A difference between their approach and ours is that they are limited to data that can be augmented, e.g., images or text. Further, they create a $k$-means like ensemble by using the Softmax, see \cite{SoftmaxAndKmeans} for the connection between the Softmax and $k$-means. In contrast to that, our \algname{} algorithm can be used with a wide range of existing clustering methods and is not limited to $k$-means. 
Liu\etal\cite{iec_LiuSLF16} proposed the IEC algorithm which embeds multiple clustering results with a marginalized Denoising \gls{ae}\cite{mDAE_ChenWSB14} and clusters the learned embedding with $k$-means, without considering the original data. Tao\etal\cite{agae_TaoLLW019} extended the idea of \cite{iec_LiuSLF16} and proposed the AGAE method. Instead of embedding the clustering results, AGAE uses a consensus graph constructed from the \gls{ca} matrix of the base partitions as an input to a \gls{dc} method, which together with the original data produces an enriched embedding. In contrast to our approach, AGAE does not learn a consensus with the neural network but uses initial clusterings to construct a consensus graph as input for their \gls{dc} algorithm, without updating the graph during training. Importantly and in contrast to ConCURL\cite{RegattiConsensus21} and \algname{}, both IEC and AGAE are not jointly updating the consensus clusterings and representation, a key feature of \gls{dc} \cite{deepclustering_survey_aljalbout18} to improve cluster performance. 

\section{Experiments}
\label{sec:experiments}
We evaluate our \algname{} algorithm with respect to several aspects. In Section \ref{sec:algorithm_analysis}, we evaluate \algname{} w.r.t. its most important hyperparameters for MNIST \cite{mnist} as it is usually done in \gls{dc} \cite{DEC,IDEC,Vade,DCN_YangFSH17} and show that our objective increases the agreement and cluster performance for all ensemble members across data sets. Additionally, we perform an ablation study across data sets. In Section~\ref{sec:cluster_performance}, we compare \algname{} to several \gls{cc} and \gls{dc} methods.

\noindent\textbf{Evaluation Metrics:} We evaluate the performance using \glsfirst{nmi} \cite{NMI} and \glsfirst{ari} \cite{ARI}. Both range between 0 and 1, where 0 indicates no match and 1 a perfect match with the ground-truth. We evaluate the agreement within an ensemble by calculating the average pairwise \gls{nmi} \cite{strehl_cluster_2002} between all clusterings. 

\noindent\textbf{Data sets:} The synthetic data set (SYNTH) consists of four clusters and is depicted in Fig. \ref{fig:synthetic_dataset}. The real-world data sets consist of commonly used \gls{dc} image data sets like MNIST, Fashion-MNIST (FMNIST) \cite{fashion_mnist}, Kuzushiji-MNIST (KMNIST) \cite{kmnist}, and USPS \cite{usps_Hull94} and three UCI \cite{UCI} data sets PENDIGITS, HAR and MICE. We provide a detailed description of the data sets in Appendix \ref{app:data_sets}. All data sets are preprocessed using a z-transformation.

\noindent\textbf{Experimental Setup:} For all data sets that have more than 2,000 data points, we use a feed-forward \gls{ae} architecture with layers $D$-500-500-2000-10 for the encoder and a corresponding mirrored decoder, which is the same setting as used in \cite{DEC}. For the MICE and SYNTH data sets, we have used smaller networks, with $D$-256-128-64 and $D$-20-20-2 for the encoders and mirrored decoders respectively. We use these architectures for \algname{}, DEC, IDEC, DCN, and VaDE. For SpectralNet\footnote{\url{https://github.com/KlugerLab/SpectralNet}} and ConCURL\footnote{\url{https://github.com/JayanthRR/ConCURL\_NCE}}, we used the settings that are available in their public implementations. IEC and AGAE have no publicly available code, which is why we only show the NMI results reported in their papers\footnote{Symbol ${\text{†}}$ indicates results are taken from \cite{iec_LiuSLF16} and ${\text{‡}}$ from \cite{agae_TaoLLW019}}.
   
   \begin{table*}[t]\caption{Ablation study for combinations of loss terms of \algname{}. Best results are marked as bold and runner-up is underlined. All results are given in NMI as mean $\pm$ std over 10 runs.}
	\centering
	\begin{tabular}{c|c|c  |c|c|c|c|c|c|c|c}
\toprule
        \textbf{$\mathcal{L}_{\text{cons}}$} & \textbf{$\mathcal{L}_{\text{CE}}$} & \textbf{$\mathcal{L}_{\text{rec}}$} & \textbf{SYNTH} & \textbf{MICE} & \textbf{PENDIGITS} & \textbf{HAR} & \textbf{MNIST} & \textbf{FMNIST} & \textbf{KMNIST} & \textbf{USPS} \\
\midrule
        X& &  & $ 0.17 \pm 0.21 $ & $ 0.44 \pm 0.05 $ & $ 0.12 \pm 0.07 $ & $ 0.01 \pm 0.01 $ & $ 0.45 \pm 0.03 $ & $ 0.42 \pm 0.02 $ & $ 0.27 \pm 0.01 $ & $ 0.52 \pm 0.04 $ \\
         &X&  & $ 0.79 \pm 0.08 $ & $ 0.42 \pm 0.02 $ & $ 0.72 \pm 0.02 $ & $ 0.68 \pm 0.04 $ & $ 0.81 \pm 0.04 $ & $ 0.58 \pm 0.01 $ & $ 0.54 \pm 0.01 $ & $ 0.77 \pm 0.01 $ \\
         & &X & $ 0.53 \pm 0.03 $ & $ 0.43 \pm 0.02 $ & $ 0.68 \pm 0.01 $ & $ 0.55 \pm 0.05 $ & $ 0.75 \pm 0.01 $ & $ 0.61 \pm 0.01 $ & $ 0.49 \pm 0.01 $ & $ 0.67 \pm 0.02 $ \\ \midrule
        X && X  & $ 0.41 \pm 0.02 $ & $ 0.50 \pm 0.04 $ & $ \underline{0.74} \pm 0.02 $ & $ 0.60 \pm 0.03 $ & $ 0.79 \pm 0.01 $ & $ \underline{0.64} \pm 0.01 $ & $ 0.52 \pm 0.02 $ & $ 0.71 \pm 0.01 $ \\ 
         &X&X & $ \underline{0.82} \pm 0.09 $ & $ 0.43 \pm 0.04 $ & $ 0.73 \pm 0.02 $ & $ 0.65 \pm 0.06 $ & $ 0.83 \pm 0.02 $ & $ 0.63 \pm 0.02 $ & $ 0.56 \pm 0.01 $ & $ 0.79 \pm 0.01 $ \\
        X&X&  & $ \textbf{0.99} \pm 0.01 $ & $ \underline{0.55} \pm 0.03 $ & $ \textbf{0.82} \pm 0.02 $ & $ \underline{0.73} \pm 0.03 $ & $ \underline{0.87} \pm 0.02 $ & $ \underline{0.64} \pm 0.01 $ & $ \underline{0.60} \pm 0.01 $ & $ \underline{0.84} \pm 0.02 $ \\ \midrule
        X&X&X & $ \textbf{0.99} \pm 0.02 $ & $ \textbf{0.57} \pm 0.03 $ & $ \textbf{0.82} \pm 0.02 $ & $ \textbf{0.75} \pm 0.02 $ & $ \textbf{0.88} \pm 0.02 $ & $ \textbf{0.65} \pm 0.01 $ & $ \textbf{0.61} \pm 0.01 $ & $ \textbf{0.85} \pm 0.01 $ \\
\bottomrule
	\end{tabular}   
	\label{tbl:ablation_study}
\end{table*}  

For hyperparameters that are specific to \algname{}, we set $\lambda_{cons}$ to $0.1$ for the image data sets and to $10.0$ for the UCI and SYNTH data sets, where the higher weight leads to better results for all data sets.
The sampling size $n$ is set to $0.08\cdot N$ for data sets with $N>11\text{,}000$ and to $0.5\cdot N$ for the others. We let our algorithm run for $T=10$ rounds and report the result with the highest agreement between ensemble members, thus not needing to specify $\tau$. We train the classifiers and encoder of \algname{} with mini-batch SGD and momentum \cite{sgd_momentum} set to $0.9$ for all data sets and $|\mathcal{B}|=256$. The classifiers are pretrained with a learning rate of $0.01$ and the representation updates are done with a learning rate of $0.001$, which is reduced by a factor of $0.9$ after each round $t$. We set the number of maximum mini-batch iterations to $\text{ITER}=20\text{,}000$ for all data sets. We used the early stopping heuristic during the classifier pretraining and consensus representation learning and decreased the learning rate by 0.9 when a loss plateau was reached. 

DEC, IDEC, and DCN are learning $k$-means friendly embeddings, SpectralNet extends spectral clustering, DeepECT learns hierarchical embeddings, and VaDE is a deep version of Gaussian mixture models. We, therefore, choose a heterogeneous ensemble of $k$-means (KM), spectral clustering (SC), agglomerative clustering (AGG), and Gaussian mixture models (GMM), based on the correspondence of the chosen \gls{dc} methods, i.e., $\ensemble=\{\text{KM}, \text{SC}, \text{AGG}, \text{GMM}\}$.

For the \gls{cc} approaches, we compare against eight methods (six classical methods, two utilizing RPs). We evaluate the \gls{cc} methods on the raw and the \gls{ae} embedded data sets using the same ensemble $\ensemble$ as \algname{}. For all methods, we assume the number of clusters $k$ to be known. 
We provide hyperparameter settings and further details for all methods in Appendix \ref{app:experiment_setting}. We uploaded the used data sets, our code and further results at \url{https://gitlab.cs.univie.ac.at/lukas/deccs}.

\begin{figure}[t]
\centering
\begin{subfigure}[t]{.465\linewidth}
\label{fig:agreement_plot_mnist}
	\includegraphics[width=\linewidth]{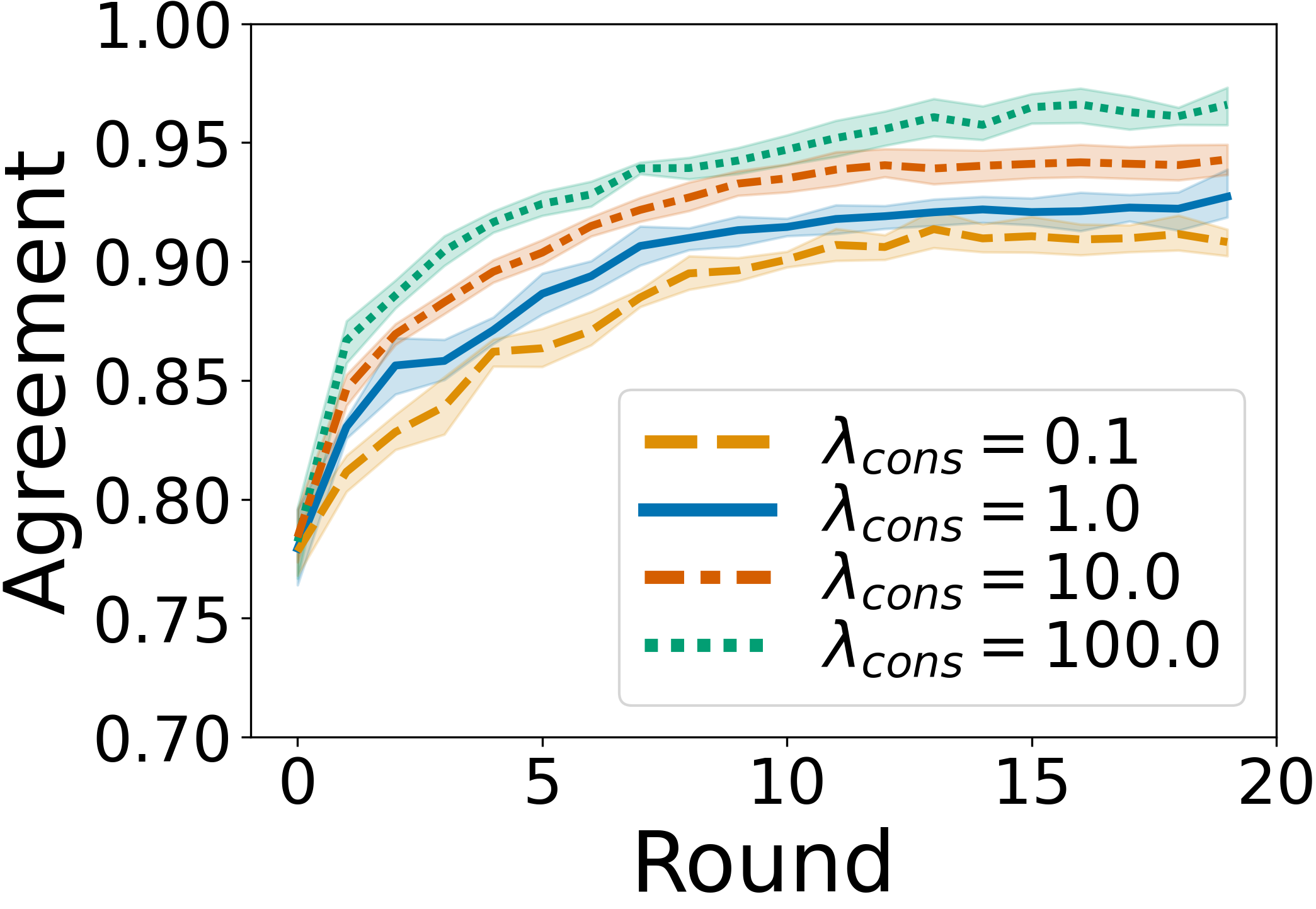}
\end{subfigure}
\hfill
\begin{subfigure}[t]{.465\linewidth}
\label{fig:performance_plot_mnist}
	\includegraphics[width=\linewidth]{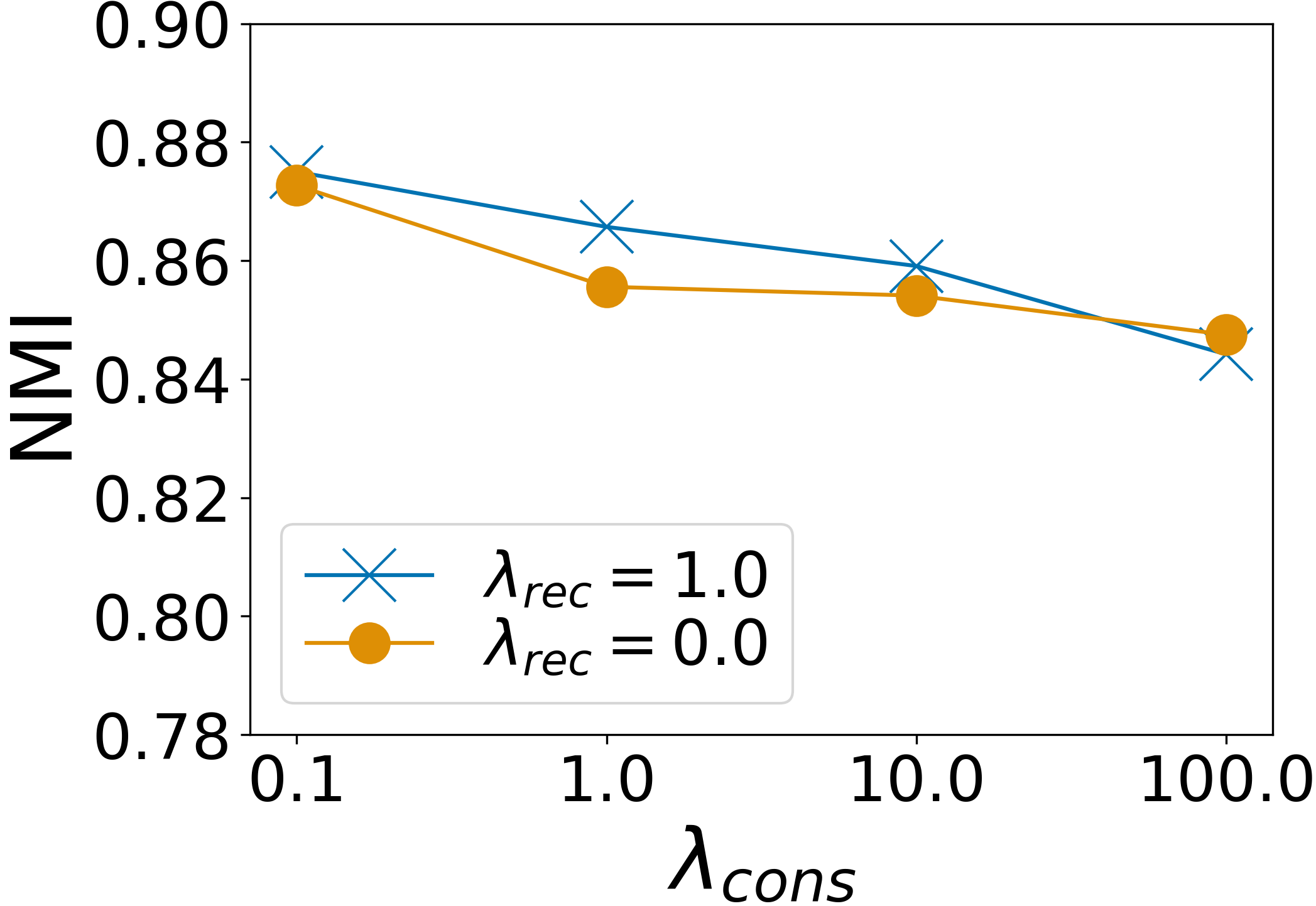}
\end{subfigure}
\caption{\algname{} parameter analysis for MNIST. (Left) Average agreement (thick lines) and 95\% confidence intervals for ten runs of \algname{} show that increasing the consensus weight $\lambda_{cons}$ leads to an increased agreement between ensemble members during training. (Right) Average cluster performance for different values of $\lambda_{cons}$ and $\lambda_{rec}$ over ten runs.}
\label{fig:parameter_analysis}
\end{figure}

\subsection{Algorithm Evaluation}
\label{sec:algorithm_analysis} 

\noindent\textbf{Ablation study}: We evaluate the impact of the individual components of \algname{}' loss function in Table \ref{tbl:ablation_study}. We see that the combination of consensus loss ($\mathcal{L}_\text{cons}$) and cross-entropy loss ($\mathcal{L}_\text{CE}$), with and without reconstruction loss ($\mathcal{L}_\text{rec}$) perform best (last two rows) for all data sets. Using only $\mathcal{L}_\text{cons}$ without $\mathcal{L}_\text{CE}$ leads to worse results because the classifiers are not preventing the merging of clusters (first row), as we have discussed in Fig. \ref{fig:consensus_update_toy}. 

\begin{figure}[t]
\centering
\begin{subfigure}[t]{.45\linewidth}
	\includegraphics[width=\linewidth]{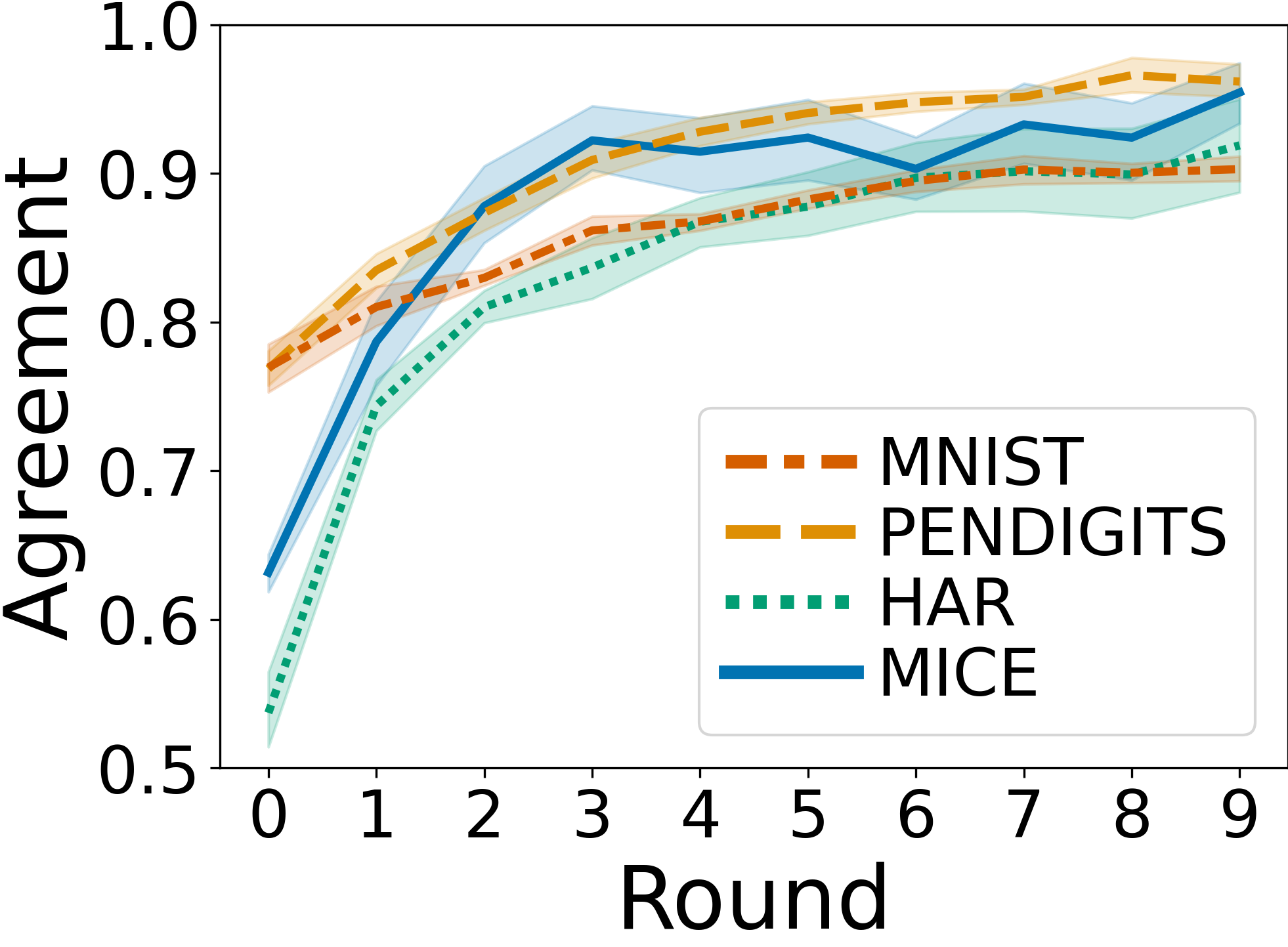}
    \label{fig:agreement_plot_others}
\end{subfigure}
\hfill
\begin{subfigure}[t]{.45\linewidth}
	\includegraphics[width=\linewidth]{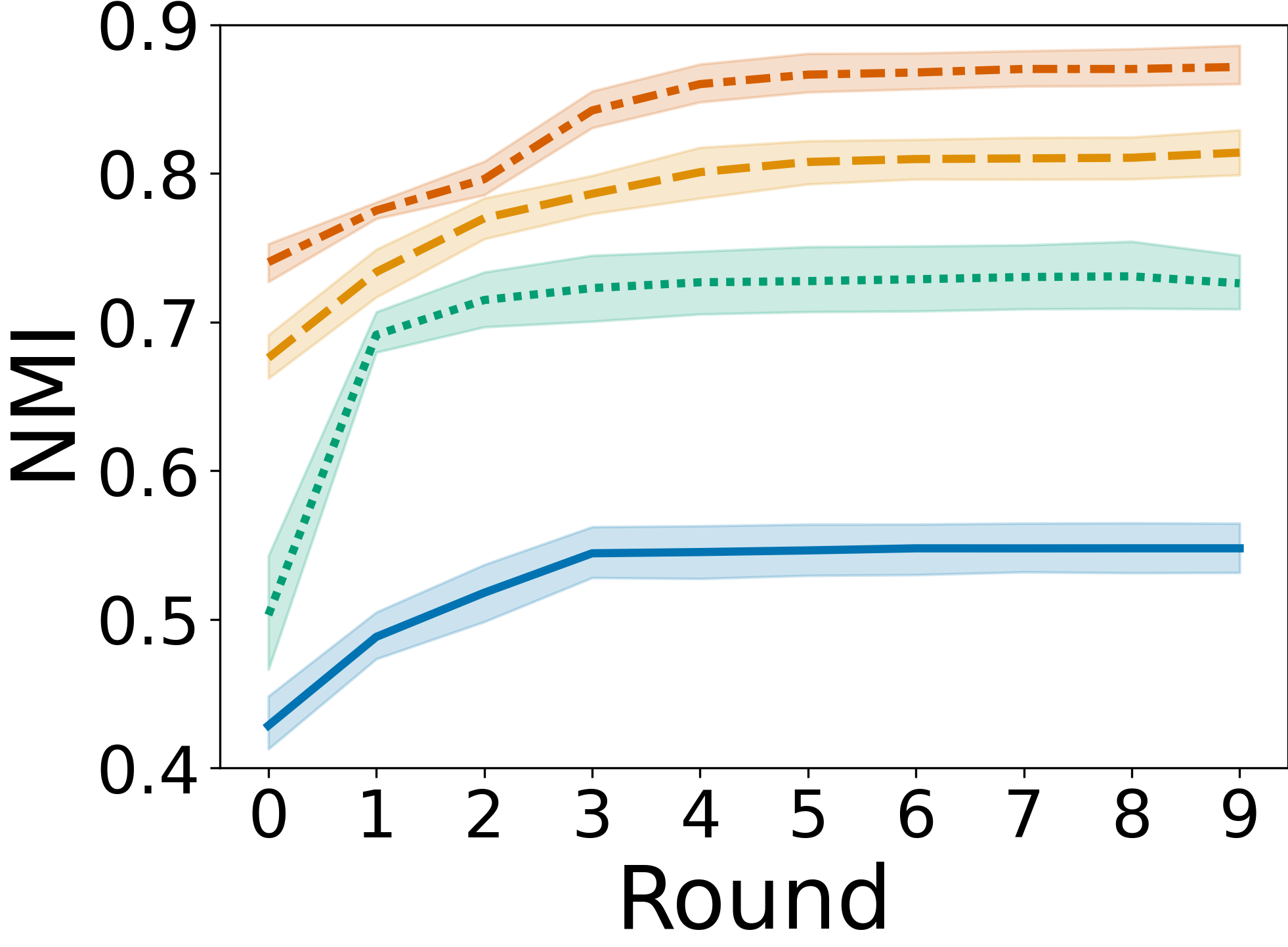}
    \label{fig:performance_plot_others}
\end{subfigure}
\vspace{-0.5cm}
\caption{Average agreement (thick lines) and 95\% confidence intervals for ten runs of \algname{} show the increase in agreement between ensemble members over training across data sets on the left side and the corresponding increase in cluster performance of \algname{} on the right side.}
\label{fig:parameter_analysis_other}
\end{figure}

\noindent\textbf{Impact of ensemble size}: We evaluated the impact of the ensemble size by increasing its original size $|\ensemble|=4$ by doubling and tripling each member in the ensemble $\ensemble$, leading to $|\ensemble_{\times 2}|=8$ and $|\ensemble_{\times 3}|=16$. We evaluated the results by averaging the cluster performance of \algname{} over ten runs on MNIST, where we achieved the same average \gls{nmi} of $0.87$ for each ensemble. We believe that we cannot see a benefit here because we use strong clustering methods, which do not add more diversity to the ensemble. 

\noindent\textbf{Impact of sampling size $n$}: To evaluate the effect of the sampling size $n$, we varied it for ratios $\{$0.02, 0.04, 0.06, 0.08$\}$ of MNIST  ($N=70\text{,}000$) and averaged the performance over ten runs. \algname{} achieved the same performance ($\text{\gls{nmi}}=0.87$) for all ratios. We chose 0.08 for the remaining experiments as this was also stable for the other large data sets.

\noindent\textbf{Impact of $\lambda_{cons}$ and $\lambda_{rec}$:} To demonstrate that the objective of \algname{} increases the agreement between cluster ensemble members, we vary the consensus weight $\lambda_{cons}$ for values in $\{$0.1, 1.0, 10.0, 100.0$\}$ for the MNIST data set while keeping $\lambda_{rec}=0$. We see on the left side of Fig. \ref{fig:parameter_analysis} that a higher value for $\lambda_{cons}$ leads to a corresponding higher agreement. This is expected because we enforce the consensus with a higher weight. The right side of Fig. \ref{fig:parameter_analysis} shows the corresponding average cluster performance for $\lambda_{rec} \in \{0.0\text{,}1.0\}$. The trend with and without the reconstruction loss is similarly downwards trending for very high values because a very highly weighted consensus loss disregards the underlying structure of the data.

\newcommand{\vpad}{2}

\begin{table*}\caption{Cluster performance results measured in NMI.
    Corresponding ARI values are shown in Table\ref{tbl:ari_results}.\label{tab:experiments}
    }
\centering

    \begin{tabular}{ l | c | c | c | c | c | c | c | c }
        \specialrule{1pt}{\vpad pt}{\vpad pt}
        \textbf{Method}
        & \textbf{SYNTH}
        & \textbf{MICE}
        & \textbf{PENDIGITS}
        & \textbf{HAR}
        & \textbf{MNIST}
        & \textbf{FMNIST}
        & \textbf{KMNIST}
        & \textbf{USPS}
\\
\cmidrule{1-9}
DECCS  &  $\textbf{0.99} \pm 0.02$  &  $\textbf{0.57} \pm 0.03$  &  $\textbf{0.82} \pm 0.02$  &  $\textbf{0.75} \pm 0.02$  &  $\underline{0.88} \pm 0.02$  &  $\textbf{0.65} \pm 0.01$  &  $\textbf{0.61} \pm 0.01$  &  $\textbf{0.85} \pm 0.01$ \\
\cmidrule{1-9}
CSPA \cite{strehl_cluster_2002}  &  $0.75 \pm 0.00$  &  $0.35 \pm 0.02$  &  $0.73 \pm 0.02$  &  $0.49 \pm 0.02$  &  $0.59 \pm 0.04$  &  $0.53 \pm 0.03$  &  $0.46 \pm 0.02$  &  $0.65 \pm 0.02$ \\
HGPA \cite{strehl_cluster_2002}  &  $0.75 \pm 0.00$  &  $0.37 \pm 0.01$  &  $0.61 \pm 0.05$  &  $0.49 \pm 0.04$  &  $0.47 \pm 0.02$  &  $0.46 \pm 0.02$  &  $0.35 \pm 0.02$  &  $0.56 \pm 0.03$ \\
MCLA \cite{strehl_cluster_2002}  &  $0.57 \pm 0.08$  &  $0.34 \pm 0.01$  &  $0.73 \pm 0.04$  &  $0.59 \pm 0.03$  &  $0.55 \pm 0.05$  &  $0.53 \pm 0.02$  &  $0.49 \pm 0.02$  &  $0.60 \pm 0.08$ \\
HBGF \cite{fern_solving_2004}  &  $0.62 \pm 0.02$  &  $0.33 \pm 0.03$  &  $0.72 \pm 0.03$  &  $0.48 \pm 0.03$  &  $0.58 \pm 0.03$  &  $0.51 \pm 0.04$  &  $0.46 \pm 0.02$  &  $0.64 \pm 0.01$ \\
NMF \cite{li_solving_2007}  &  $0.46 \pm 0.05$  &  $0.34 \pm 0.01$  &  $0.75 \pm 0.03$  &  $0.59 \pm 0.03$  &  $0.59 \pm 0.03$  &  $0.52 \pm 0.03$  &  $0.49 \pm 0.03$  &  $0.72 \pm 0.03$ \\
LWEA \cite{huang_locally_2018}  &  $0.60 \pm 0.02$  &  $0.37 \pm 0.03$  &  $\underline{0.76} \pm 0.02$  &  $0.59 \pm 0.00$  &  $0.62 \pm 0.03$  &  $0.56 \pm 0.01$  &  $0.51 \pm 0.01$  &  $0.72 \pm 0.01$ \\
RP+EM \cite{fern_random_2003}  &  $0.61 \pm 0.00$  &  $0.46 \pm 0.04$  &  $0.67 \pm 0.05$  &  $0.46 \pm 0.06$  &  $0.48 \pm 0.05$  &  $0.50 \pm 0.05$  &  $0.44 \pm 0.04$  &  $0.64 \pm 0.04$ \\
RP+FCM \cite{popescu_random_2015}  &  $0.64 \pm 0.05$  &  $0.28 \pm 0.08$  &  $0.63 \pm 0.01$  &  $0.51 \pm 0.01$  &  $0.22 \pm 0.02$  &  $0.40 \pm 0.01$  &  $0.25 \pm 0.01$  &  $0.38 \pm 0.02$ \\
\cmidrule{1-9}
AE+CSPA \cite{strehl_cluster_2002}  &  $\underline{0.76} \pm 0.05$  &  $0.41 \pm 0.03$  &  $0.73 \pm 0.02$  &  $0.53 \pm 0.01$  &  $0.84 \pm 0.02$  &  $0.59 \pm 0.02$  &  $0.54 \pm 0.02$  &  $0.74 \pm 0.03$ \\
AE+HGPA \cite{strehl_cluster_2002}  &  $0.75 \pm 0.00$  &  $0.41 \pm 0.03$  &  $0.64 \pm 0.07$  &  $0.49 \pm 0.05$  &  $0.61 \pm 0.02$  &  $0.50 \pm 0.03$  &  $0.43 \pm 0.03$  &  $0.60 \pm 0.02$ \\
AE+MCLA \cite{strehl_cluster_2002}  &  $0.53 \pm 0.11$  &  $0.43 \pm 0.02$  &  $0.73 \pm 0.05$  &  $0.59 \pm 0.06$  &  $0.83 \pm 0.01$  &  $0.62 \pm 0.03$  &  $\underline{0.59} \pm 0.02$  &  $0.75 \pm 0.06$ \\
AE+HBGF \cite{fern_solving_2004}  &  $0.65 \pm 0.08$  &  $0.40 \pm 0.03$  &  $0.71 \pm 0.04$  &  $0.53 \pm 0.03$  &  $0.83 \pm 0.02$  &  $0.58 \pm 0.02$  &  $0.54 \pm 0.02$  &  $0.72 \pm 0.02$ \\
AE+NMF \cite{li_solving_2007}  &  $0.50 \pm 0.13$  &  $0.44 \pm 0.04$  &  $\underline{0.76} \pm 0.04$  &  $0.60 \pm 0.01$  &  $0.82 \pm 0.04$  &  $0.61 \pm 0.02$  &  $\underline{0.59} \pm 0.03$  &  $0.82 \pm 0.04$ \\
AE+LWEA \cite{huang_locally_2018}  &  $0.61 \pm 0.04$  &  $0.46 \pm 0.04$  &  $0.75 \pm 0.03$  &  $0.58 \pm 0.06$  &  $0.86 \pm 0.02$  &  $\textbf{0.65} \pm 0.01$  &  $\textbf{0.61} \pm 0.03$  &  $\underline{0.83} \pm 0.03$ \\
AE+RP+EM \cite{fern_random_2003}  &  $0.62 \pm 0.03$  &  $\underline{0.51} \pm 0.04$  &  $0.67 \pm 0.04$  &  $0.48 \pm 0.04$  &  $0.77 \pm 0.04$  &  $0.59 \pm 0.02$  &  $0.58 \pm 0.03$  &  $0.65 \pm 0.03$ \\
AE+RP+FCM \cite{popescu_random_2015}  &  $0.63 \pm 0.10$  &  $0.41 \pm 0.03$  &  $0.54 \pm 0.05$  &  $0.48 \pm 0.03$  &  $0.45 \pm 0.07$  &  $0.49 \pm 0.03$  &  $0.31 \pm 0.04$  &  $0.37 \pm 0.03$ \\
\cmidrule{1-9}
SpectralNet \cite{SpectralNet18}  &  $0.72 \pm 0.06$  &  $0.27 \pm 0.06$  &  $\textbf{0.82} \pm 0.04$  &  $\underline{0.61} \pm 0.06$  &  $\textbf{0.92} \pm 0.00$  &  $\underline{0.64} \pm 0.01$  &  $\textbf{0.61} \pm 0.02$  &  $\underline{0.83} \pm 0.02$ \\
DEC \cite{DEC}  &  $0.65 \pm 0.03$  &  $0.49 \pm 0.02$  &  $0.75 \pm 0.02$  &  $0.54 \pm 0.09$  &  $0.84 \pm 0.01$  &  $0.60 \pm 0.01$  &  $0.52 \pm 0.01$  &  $0.80 \pm 0.01$ \\
IDEC \cite{IDEC}  &  $0.64 \pm 0.03$  &  $0.50 \pm 0.03$  &  $\underline{0.76} \pm 0.02$  &  $0.53 \pm 0.09$  &  $0.85 \pm 0.02$  &  $0.62 \pm 0.02$  &  $0.55 \pm 0.03$  &  $0.81 \pm 0.01$ \\
DCN \cite{DCN_YangFSH17}  &  $0.59 \pm 0.08$  &  $0.48 \pm 0.04$  &  $0.75 \pm 0.02$  &  $0.51 \pm 0.08$  &  $0.84 \pm 0.03$  &  $0.62 \pm 0.02$  &  $0.54 \pm 0.04$  &  $0.78 \pm 0.04$ \\
VaDE \cite{VAE}  &  $0.62 \pm 0.10$  &  $0.45 \pm 0.06$  &  $0.75 \pm 0.02$  &  $0.54 \pm 0.09$  &  $0.83 \pm 0.03$  &  $\textbf{0.65} \pm 0.01$  &  $0.56 \pm 0.01$  &  $0.79 \pm 0.03$ \\
DeepECT \cite{ECT}  &  $0.61 \pm 0.10$  &  $0.47 \pm 0.06$  &  $0.74 \pm 0.02$  &  $0.56 \pm 0.10$  &  $0.82 \pm 0.03$  &  $0.62 \pm 0.03$  &  $0.52 \pm 0.04$  &  $0.76 \pm 0.06$ \\
ConCURL \cite{RegattiConsensus21}  &  n.a.  &  n.a.  &  n.a.  &  n.a.  &  $0.60 \pm 0.04$  &  $0.48 \pm 0.02$  &  $0.30 \pm 0.03$  &  $0.49 \pm 0.02$ \\
IEC \cite{iec_LiuSLF16}  &  -  &  -  &  $0.72^{\text{‡}}$  &  -  &  $0.54^{\text{†}}$  &  -  &  -  &  $0.64^{\text{†}}$ \\
AGAE \cite{agae_TaoLLW019}  &  -  &  -  &  $0.74^{\text{‡}}$  &  -  &  -  &  -  &  -  &  $0.74^{\text{‡}}$ \\

        \specialrule{1pt}{\vpad pt}{\vpad pt}
    \end{tabular}

\end{table*}

\noindent\textbf{Increase of agreement and \gls{nmi} during training:} On the left side of Fig. \ref{fig:parameter_analysis_other}, we show how \algname{} increases the ensemble agreement over training for three UCI data sets respectively, and for MNIST as the behavior for the image data sets was very similar. This experiment gives additional evidence that our algorithm can effectively maximize the pairwise \gls{nmi}  between ensemble members (Eq. \ref{eq:min_objective_nmi}) by learning a consensus representation. The agreement is stabilizing for all data sets at round 8, except for MICE which fluctuates at a high agreement level due to the smaller data set size. The right side of Fig. \ref{fig:parameter_analysis_other} shows the corresponding increase in cluster performance. We see that \algname{} reaches stable cluster performance already after round five for all data sets. 

\subsection{Cluster performance}
\label{sec:cluster_performance}
In Table \ref{tab:experiments}, we show the clustering results of all methods w.r.t. \gls{nmi} over ten runs. We see in Table \ref{tab:experiments} that for the SYNTH, MICE and HAR data set \algname{} clearly outperforms the next best method. For the PENDIGITS data set, we perform similar to SpectralNet in \gls{nmi} and outperform it w.r.t. ARI (0.73 vs 0.67). For the MICE data set, we see that all \gls{cc} methods improve when applied to the \gls{ae} embedded space, but \algname{} is still outperforming them, showing that updating the representation can increase the cluster performance even further. The highest improvement for the real-world data sets can be seen for the HAR data set, where we outperform the next best clustering method (\gls{nmi}=0.61) by 0.14. The results on the image data sets show that \algname{} outperforms all comparison methods on USPS. \algname{} performs similar to the \gls{dc} methods for MNIST, FMNIST, and KMNIST. For MNIST, we are only outperformed by SpectralNet. Interestingly, the \gls{cc} algorithms that are applied to the embedded space for the image data sets serve as strong baselines, e.g., reaching 0.65 and 0.61 \gls{nmi} for FMNIST and KMNIST respectively. \algname{} outperforms the deep consensus clustering method ConCURL for all image data sets. ConCURL heavily relies on image augmentation and for small, greyscale images there are fewer augmentation invariances available, which might be the reason for ConCurl's poor performance. Further, ConCURL cannot be applied to non-augmentable data, which is why these results are marked as not applicable (n.a.).

\section{Discussion and Conclusion}

\noindent\textbf{Noise and outlier points}: Currently, we have not considered noise-aware clustering methods, like DBSCAN \cite{dbscan}, in our ensembles. \algname{} could be extended to include methods like DBSCAN, e.g., by excluding noise and outlier points during the representation update, such that a consensus representation is learned only for inlier clusters. 

\noindent\textbf{Consensus representation learning}: With \algname{} we have introduced the first algorithm to learn consensus representations for \gls{cc}. In future work, we would like to explore alternative approaches for optimizing the proposed objective in Eq. \ref{eq:min_objective_nmi}, which could lead to novel approaches to \gls{cc}.

We have proposed the idea of consensus representations, a novel way of learning a \gls{cc} by maximizing the agreement between ensemble members using representation learning. Additionally, we have introduced the \algname{} algorithm, to the best of our knowledge, it is the first \gls{dc} algorithm that can use multiple heterogeneous clustering methods to jointly improve the learned representation and clustering results.

\bibliographystyle{IEEEtran}
\bibliography{literature}

\clearpage
\appendix

\subsection{DATA SETS}
\label{app:data_sets}

\noindent\textbf{Mice Protein Expression (MICE)} \cite{UCI}: Data set consisting of 552 vectors with  77 dimensions and  8 ground-truth clusters. Each vector represents the expression levels of 77 proteins of the mice's cortex. 

\noindent\textbf{Pendigits} \cite{UCI}:
Data set consisting of 10,992 vectors with 16 dimensions, representing 8 coordinates. The coordinates were gathered during the writing of digits (0 to 9) on a tablet.

\noindent\textbf{Human Activity Recognition (HAR)} \cite{UCI}:
Data set consisting of 10,299 vectors with 561 dimensions with records from smartphones and smartwatches. The data set contains six clusters corresponding to different human activities.

\noindent\textbf{MNIST} \cite{mnist}: 
Data set consisting of 70,000 hand-written digits (0 to 9) with a size of $28 \times 28$ pixels.

\noindent\textbf{FMNIST} \cite{fashion_mnist}:
Data set consisting of 70,000 goods from the Zalando online store. Each sample belongs to one of 10 products and has a size of $28 \times 28$ pixels.

\noindent\textbf{KMNIST} \cite{kmnist}: 
Data set consisting of 70,000 Kanji characters (10 different characters) with a size of $28 \times 28$ pixels.

\noindent\textbf{USPS} \cite{usps_Hull94}:
Data set consisting of 9,298 hand-written digits (0 to 9) with a size of $16 \times 16$ pixels.

\begin{figure}[th]
    \centering
    \includegraphics[width=0.9\linewidth]{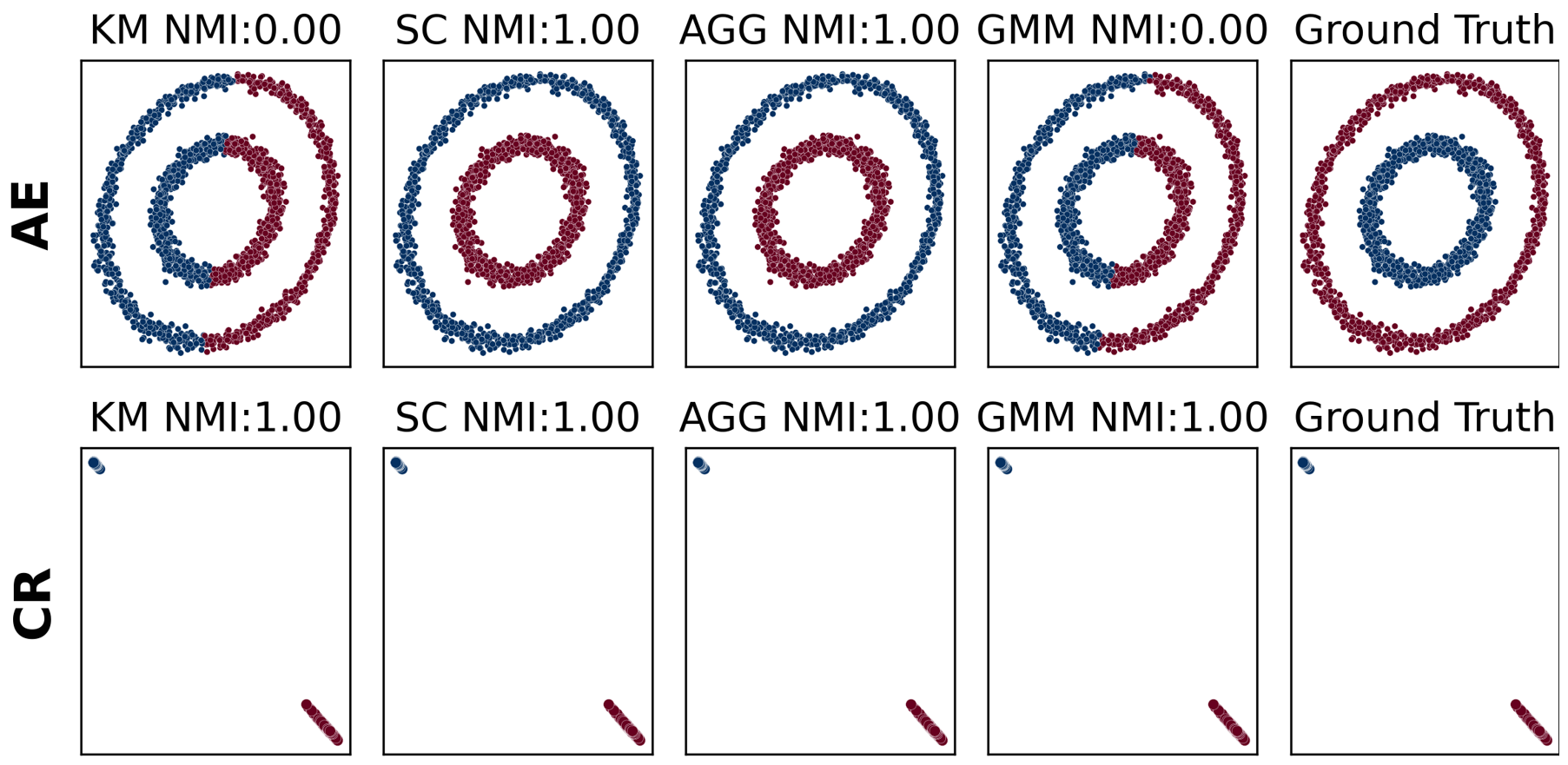}
    \caption{Consensus representation (CR) learned with \algname{} for two not linearly separable clusters. \textit{Upper row}: Initial clustering results on an \glsfirst{ae} embedding. Two ensemble members (SC, AGG) can perfectly discover the ground-truth clustering, but the two others (KM, GMM) do not perform better than chance. \textit{Lower row}: Clustering results on a CR learned with \algname{} leads to the same, perfect performance for all clustering algorithms}
    \label{fig:non-linear-cluster}
\end{figure}

\subsection{EXPERIMENT SETUP}
\label{app:experiment_setting}
\noindent\textbf{Hardware setup:} We trained all \gls{dc} algorithms on a machine with a single NVIDIA RTX 2080TI GPU (11GB onboard memory), 96 GB RAM, and an Intel(R) Xeon(R) Gold 6130 CPU. All other comparison methods were run on the same machine using only the CPU.

\noindent\textbf{Implementation:} We implemented \algname{} in PyTorch (\url{https://pytorch.org/}). Currently, the generation and approximation of base partitions is done sequentially, but could be further optimized using parallelization. One run of \algname{} for MNIST took about one hour, which is in the same order of magnitude as other \gls{dc} methods, e.g. DCN needed about 40 minutes.

\noindent\textbf{Parameters of cluster ensemble:}
For the clustering algorithms in the ensemble $\ensemble=\{$KM, SC, AGG, GMM$\}$, we used the implementations of the \textit{sklearn} \cite{scikit-learn} package. For the real-world data sets, we parameterized them using this setting:

\begin{lstlisting}[language=Python]
KMeans(n_clusters=k)

SpectralClustering(n_clusters=k, 
                   affinity='nearest_neighbors',
                   n_neighbors=10, 
                   assign_labels='kmeans')

AgglomerativeClustering(n_clusters=k, 
                        linkage='ward')

GaussianMixture(n_components=k, 
                covariance_type='full',
                reg_covar=1e-5)
\end{lstlisting}
For the synthetic data sets, we set $\text{linkage}=\text{'single'}$ for agglomerative clustering.

\noindent\textbf{Consensus Clustering}: For the classical methods, we used the same parameterization as suggested by the authors in the original papers (\cite{strehl_cluster_2002}, \cite{li_solving_2007}, \cite{fern_solving_2004}, \cite{huang_locally_2018}). For the RP-based methods, we determined the subspace dimension with a grid search and report the value with the best average \gls{nmi}  over 10 runs. We evaluated the RP algorithms with an ensemble size of four (size of our ensemble) and 30 (ensemble size used in \cite{popescu_random_2015}) and again picked the run with the highest \gls{nmi}. The parameters of the \gls{em} and \gls{fcm} algorithms were set according to \cite{fern_random_2003} and \cite{popescu_random_2015} respectively.  \cite{popescu_random_2015} proposed two versions of their RP-based algorithm, we run our experiments with both versions and only report the run with the highest \gls{nmi}. 
Aligned with \cite{popescu_random_2015}, we performed a grid search for q in [5, 100] for both RP-based methods and picked the run with the highest \gls{nmi}. For \gls{lwea}, we set the hyperparameter $\theta$ to 0.4, as \cite{huang_locally_2018} do for all their experiments. Except for \gls{lwea}, other classical methods are implemented using the cluster ensemble package.\footnote{\url{https://github.com/827916600/ClusterEnsembles}}

\noindent\textbf{Deep Clustering}:
For the UCI data sets, we determined the learning rate and other \gls{ae} parameters, like dropout rate \cite{dropout}, using a grid search during the \gls{ae} pretraining taking the parameters with lowest reconstruction loss. All other \gls{ae}s were pretrained with a learning rate of $0.001$. All \gls{ae}s were trained using early stopping and the learning rate was reduced by 0.5 if the reconstruction loss reached a plateau. We use for \algname{}, VaDE, DEC, IDEC, and DCN a batch size of $|\mathcal{B}|=256$. For VaDE, DEC, IDEC, and DCN, we use a constant learning rate of $0.0001$ for the joint clustering. We train the \gls{dc} algorithms for 100,000 mini-batch iterations for all data sets using the Adam optimizer \cite{adam} as was done in the original papers. For SpectralNet \cite{SpectralNet18}, we used for the SYNTH and MICE data set the same parameters as suggested for small data sets. For the other data sets (PENDIGITS, HAR, and image data), we used the same parameters as \cite{SpectralNet18} used for MNIST.

\begin{table*}\caption{Cluster performance results measured in ARI. Best results are marked as bold and runner-up is underlined. All results are given in ARI as mean $\pm$ std over 10 runs. \algname{} $\lambda_\text{rec}=1$ indicates that $\mathcal{L}_\text{rec}$ was used, while $\lambda_\text{rec}=0$ are the results without $\mathcal{L}_\text{rec}$.
    }\label{tbl:ari_results}
\centering

    \begin{tabular}{ l | c | c | c | c | c | c | c | c }
        \specialrule{1pt}{\vpad pt}{\vpad pt}
        \textbf{Method}
        & \textbf{SYNTH}
        & \textbf{MICE}
        & \textbf{PENDIGITS}
        & \textbf{HAR}
        & \textbf{MNIST}
        & \textbf{FMNIST}
        & \textbf{KMNIST}
        & \textbf{USPS}
\\
\cmidrule{1-9}
DECCS $(\lambda_{rec} = 1)$ & $\underline{0.99} \pm 0.02$ & $\textbf{0.36} \pm 0.04$ & $\textbf{0.73} \pm 0.03$ & $\textbf{0.65} \pm 0.03$ & $\underline{0.85} \pm 0.05$ & $\underline{0.47} \pm 0.01$ & $\textbf{0.45} \pm 0.01$ & $\textbf{0.78} \pm 0.01$ \\
DECCS $(\lambda_{rec} = 0)$ & $\textbf{1.00} \pm 0.01$ & $\underline{0.33} \pm 0.03$ & $\underline{0.72} \pm 0.04$ & $\underline{0.62} \pm 0.03$ & $\underline{0.85} \pm 0.04$ & $\underline{0.47} \pm 0.01$ & $\underline{0.44} \pm 0.02$ & $\underline{0.77} \pm 0.02$ \\
\cmidrule{1-9}
CSPA \cite{strehl_cluster_2002} & $0.67 \pm 0.00$ & $0.19 \pm 0.02$ & $0.64 \pm 0.04$ & $0.37 \pm 0.02$ & $0.49 \pm 0.05$ & $0.40 \pm 0.04$ & $0.36 \pm 0.02$ & $0.54 \pm 0.02$ \\
HGPA \cite{strehl_cluster_2002} & $0.67 \pm 0.00$ & $0.20 \pm 0.01$ & $0.47 \pm 0.08$ & $0.36 \pm 0.05$ & $0.34 \pm 0.03$ & $0.30 \pm 0.03$ & $0.23 \pm 0.03$ & $0.42 \pm 0.04$ \\
MCLA \cite{strehl_cluster_2002} & $0.43 \pm 0.08$ & $0.17 \pm 0.01$ & $0.61 \pm 0.06$ & $0.46 \pm 0.03$ & $0.42 \pm 0.04$ & $0.36 \pm 0.02$ & $0.35 \pm 0.02$ & $0.52 \pm 0.08$ \\
HBGF \cite{fern_solving_2004} & $0.56 \pm 0.02$ & $0.17 \pm 0.03$ & $0.64 \pm 0.03$ & $0.37 \pm 0.04$ & $0.49 \pm 0.03$ & $0.39 \pm 0.05$ & $0.37 \pm 0.03$ & $0.54 \pm 0.02$ \\
NMF \cite{li_solving_2007} & $0.37 \pm 0.04$ & $0.19 \pm 0.02$ & $0.63 \pm 0.05$ & $0.46 \pm 0.06$ & $0.47 \pm 0.05$ & $0.37 \pm 0.03$ & $0.37 \pm 0.02$ & $0.59 \pm 0.05$ \\
LWEA \cite{huang_locally_2018} & $0.44 \pm 0.02$ & $0.18 \pm 0.02$ & $0.62 \pm 0.03$ & $0.46 \pm 0.00$ & $0.50 \pm 0.05$ & $0.40 \pm 0.02$ & $0.35 \pm 0.03$ & $0.63 \pm 0.01$ \\
RP+EM \cite{fern_random_2003} & $0.44 \pm 0.00$ & $0.26 \pm 0.04$ & $0.42 \pm 0.11$ & $0.31 \pm 0.07$ & $0.22 \pm 0.04$ & $0.29 \pm 0.07$ & $0.24 \pm 0.05$ & $0.45 \pm 0.07$ \\
RP+FCM \cite{popescu_random_2015} & $0.53 \pm 0.10$ & $0.15 \pm 0.05$ & $0.47 \pm 0.02$ & $0.32 \pm 0.00$ & $0.12 \pm 0.02$ & $0.25 \pm 0.01$ & $0.14 \pm 0.01$ & $0.20 \pm 0.06$ \\
\cmidrule{1-9}
AE+CSPA \cite{strehl_cluster_2002} & $0.69 \pm 0.07$ & $0.23 \pm 0.03$ & $0.65 \pm 0.02$ & $0.40 \pm 0.02$ & $0.82 \pm 0.03$ & $0.46 \pm 0.02$ & $0.43 \pm 0.02$ & $0.64 \pm 0.04$ \\
AE+HGPA \cite{strehl_cluster_2002} & $0.67 \pm 0.01$ & $0.23 \pm 0.02$ & $0.51 \pm 0.08$ & $0.36 \pm 0.05$ & $0.47 \pm 0.04$ & $0.34 \pm 0.04$ & $0.29 \pm 0.03$ & $0.45 \pm 0.03$ \\
AE+MCLA \cite{strehl_cluster_2002} & $0.42 \pm 0.10$ & $0.23 \pm 0.03$ & $0.62 \pm 0.07$ & $0.43 \pm 0.08$ & $0.78 \pm 0.03$ & $0.46 \pm 0.02$ & $\textbf{0.45} \pm 0.03$ & $0.68 \pm 0.07$ \\
AE+HBGF \cite{fern_solving_2004} & $0.59 \pm 0.10$ & $0.22 \pm 0.03$ & $0.62 \pm 0.05$ & $0.41 \pm 0.03$ & $0.80 \pm 0.03$ & $0.44 \pm 0.02$ & $\underline{0.44} \pm 0.02$ & $0.64 \pm 0.03$ \\
AE+NMF \cite{li_solving_2007} & $0.41 \pm 0.15$ & $0.25 \pm 0.04$ & $0.66 \pm 0.07$ & $0.41 \pm 0.03$ & $0.75 \pm 0.08$ & $\textbf{0.48} \pm 0.03$ & $\underline{0.44} \pm 0.05$ & $0.74 \pm 0.09$ \\
AE+LWEA \cite{huang_locally_2018} & $0.45 \pm 0.04$ & $0.25 \pm 0.04$ & $0.62 \pm 0.05$ & $0.41 \pm 0.09$ & $0.81 \pm 0.02$ & $\underline{0.47} \pm 0.01$ & $\textbf{0.45} \pm 0.05$ & $\underline{0.77} \pm 0.06$ \\
AE+RP+EM \cite{fern_random_2003} & $0.46 \pm 0.04$ & $0.31 \pm 0.05$ & $0.38 \pm 0.08$ & $0.38 \pm 0.03$ & $0.68 \pm 0.09$ & $0.42 \pm 0.03$ & $0.40 \pm 0.04$ & $0.48 \pm 0.03$ \\
AE+RP+FCM \cite{popescu_random_2015} & $0.54 \pm 0.16$ & $0.22 \pm 0.03$ & $0.37 \pm 0.04$ & $0.35 \pm 0.04$ & $0.29 \pm 0.08$ & $0.30 \pm 0.03$ & $0.18 \pm 0.04$ & $0.21 \pm 0.05$ \\
\cmidrule{1-9}
SpectralNet \cite{SpectralNet18} & $0.53 \pm 0.08$ & $0.15 \pm 0.04$ & $0.67 \pm 0.08$ & $0.46 \pm 0.09$ & $\textbf{0.93} \pm 0.00$ & $\underline{0.47} \pm 0.00$ & $0.42 \pm 0.04$ & $0.67 \pm 0.05$ \\
DEC \cite{DEC} & $0.50 \pm 0.03$ & $0.27 \pm 0.03$ & $0.61 \pm 0.04$ & $0.39 \pm 0.11$ & $0.81 \pm 0.02$ & $0.44 \pm 0.02$ & $0.39 \pm 0.01$ & $0.73 \pm 0.01$ \\
IDEC \cite{IDEC} & $0.49 \pm 0.03$ & $0.29 \pm 0.03$ & $0.62 \pm 0.04$ & $0.40 \pm 0.11$ & $0.82 \pm 0.03$ & $0.46 \pm 0.03$ & $0.41 \pm 0.03$ & $0.74 \pm 0.01$ \\
DCN \cite{DCN_YangFSH17} & $0.44 \pm 0.08$ & $0.27 \pm 0.04$ & $0.60 \pm 0.04$ & $0.36 \pm 0.11$ & $0.79 \pm 0.06$ & $0.45 \pm 0.03$ & $0.38 \pm 0.05$ & $0.70 \pm 0.07$ \\
VaDE \cite{VAE} & $0.49 \pm 0.13$ & $0.25 \pm 0.05$ & $0.61 \pm 0.04$ & $0.38 \pm 0.10$ & $0.78 \pm 0.06$ & $\textbf{0.48} \pm 0.02$ & $0.40 \pm 0.02$ & $0.70 \pm 0.06$ \\
DeepECT \cite{ECT} & $0.47 \pm 0.12$ & $0.27 \pm 0.06$ & $0.60 \pm 0.04$ & $0.41 \pm 0.12$ & $0.76 \pm 0.06$ & $0.44 \pm 0.05$ & $0.36 \pm 0.05$ & $0.67 \pm 0.09$ \\
ConCURL \cite{RegattiConsensus21} & n.a. & n.a. & n.a. & n.a. & $0.48 \pm 0.05$ & $0.34 \pm 0.02$ & $0.20 \pm 0.03$ & $0.33 \pm 0.02$ \\
        \specialrule{1pt}{\vpad pt}{\vpad pt}
    \end{tabular}

\end{table*}

\noindent\textbf{ConCURL} \cite{RegattiConsensus21}:
For ConCURL, we used the author's repository. For all data sets, we performed ten runs each. In each run, we trained the algorithm for 300 epochs. To change the basic architecture as little as possible, we transformed gray images into three dimensions by copying the gray color channel. In our experiment, we used PyTorch's resnet18 with a hidden MLP of 2048. We chose SGD as the optimizer with a learning rate of 0.015 and set the batch size to 128.
The alpha parameter was set to 0, and the beta and gamma parameters were set to 1. In the experiment, we set NCE-temp to 0.085 and NCE-k to 4096. For the hyperparameters, we followed the hyperparameters of CIFAR-10 available in their repository. For image augmentation, we used random rotations between -10 and 10 degrees, translations between 0 and 0.1, scaling between 0.6 and 1.2, and shearing between -10 and 10.



\end{document}